\documentclass[letterpaper]{article} 
\usepackage{aaai24}  
\usepackage{times}  
\usepackage{helvet}  
\usepackage{courier}  
\usepackage[hyphens]{url}  
\usepackage{graphicx} 
\usepackage{natbib}  
\usepackage{caption} 
\usepackage{algorithm}
\usepackage{algorithmic}
\usepackage{nameref}
\usepackage{newfloat}
\usepackage{listings}
\DeclareCaptionStyle{ruled}{labelfont=normalfont,labelsep=colon,strut=off} 
\floatstyle{ruled}
\newfloat{listing}{tb}{lst}{}
\floatname{listing}{Listing}
\title{Rating Multi-Modal Time-Series Forecasting Models (MM-TSFM) for Robustness Through a Causal Lens}
\author{
Kausik Lakkaraju$^1$, Rachneet Kaur$^2$, Zhen Zeng$^2$, Parisa Zehtabi$^3$, \\ Sunandita Patra$^2$, 
Biplav Srivastava$^1$, Marco Valtorta$^1$
}
\affiliations{
    \textsuperscript{\rm 1}University of South Carolina, USA, \textsuperscript{\rm 2}J.P. Morgan AI Research, USA, \textsuperscript{\rm 3}J.P. Morgan AI Research, UK\\

}

\usepackage{bibentry}
\usepackage{amsmath}

\usepackage{subcaption}
\usepackage{array}
\usepackage{xcolor}   
\usepackage{multirow}

\begin{document}

\maketitle

\begin{abstract}
 AI systems are notorious for their fragility, e.g., minor input changes can potentially cause major output swings.
When such systems are deployed in critical areas like finance, the consequences of their uncertain behavior could be severe. In this paper, we focus on multi-modal time-series forecasting in which imprecision due to noisy or incorrect data can lead to erroneous predictions, impacting stakeholders, such as analysts, investors, and traders. Recently, it has been shown that beyond numeric data, its graphical transformations can be used with advanced visual models to get better performance. 
In this context, we introduce a rating methodology to assess the robustness of Multi-Modal Time-Series Forecasting Models (MM-TSFM) through causal analysis which helps us understand and quantify the isolated impact of various attributes on the forecasting accuracy of MM-TSFM.

We test these systems across three different settings: in the presence of noisy data, with erroneous data, and in conjunction with another AI system that may exhibit bias. 
We apply our novel rating method on a variety of numeric and multimodal forecasting models in a large experimental setup (six input settings of control and perturbations, ten data distributions, time series from six leading stocks in three industries over a year of data, and five time-series forecasters) to draw insights on robust forecasting models and the context of their strengths.
Within the scope of our study, our main result is that multi-modal (numeric+visual) forecasting, which was found to be more accurate than just numeric forecasting in previous studies, can also be more robust, in diverse settings. 
Our work will help different stakeholders of time series forecasting understand the models' behaviors along {\em trust (robustness)} and {\em accuracy dimensions} to select an appropriate model for forecasting using our rating method leading to improved and inclusive decision-making.  
\end{abstract}

\section{Introduction}

Artificial Intelligence (AI) systems, regardless of the mode in which they take input - numeric, textual, audio, visual, or multimodal,  are notorious for their
fragility (lack of robustness) and other characteristics (e.g., opaqueness, alignment to human values) that go beyond performance to contribute to users' trust of technology\cite{trust-ml-book}. For example, small variations in the inputs to a Machine Learning (ML) model may
result in 
drastic swings in its output. 
This uncertainty about robustness is amplified by the lack of interpretability of many ML models due to their black-box nature \cite{explainable-ai-survey}.
As a result, such systems face challenges in gaining acceptance and trust from end-users hampering their widespread adoption.

When such systems are deployed in critical areas like finance \cite{imf-ai-impact-report}
and healthcare \cite{ai-health-report}, the consequences of their uncertain behavior could cause critical failures. A promising idea to manage user trust is to {\em communicate} the behavior of AI systems through ratings that are assigned after assessing AI systems from a third-party perspective (without access to the system's training data). These methods \cite{srivastava2018towards,srivastava2020rating,srivastava2023advances-rating,kausik2023the,kausik2024rating}, have considered text-based AI systems like automated machine translators and sentiment analysis systems in various settings including multi-lingual text, using which it was shown that users can make informed decisions when a choice of AI models are available to them. Along these lines, we consider multimodal data and time-series forecasting-based AI models.


Time-series forecasting is a popular form of AI usage in many industries including finance. 
Here, imprecision due to noisy or incorrect data can lead to erroneous predictions, impacting stakeholders who rely on AI models for their day-to-day work. Many current models in this domain use multimodal AI systems to predict trends and stock prices due to their effectiveness over numerical data. \cite{zeng2023from} is one such work in which the authors introduced ViT-num-spec MM-TSFM and showed that it outperformed other numerical and image-based models.
Hence, we select daily stock price prediction task to assess various time-series forecasting models including two MM-TSFM that use ViT-num-spec architecture. We collected 13 months of stock prices from six leading companies, two from each of the three critical industries: Technology, Pharmaceuticals, and Finance. We organized this data into 2,75,400 data points
across five model predictions, ten sampled distributions, and six perturbed variations. 

Our contributions are that we:
1. Introduce a causal analysis-based method to rate Multimodal Time-Series Forecasting Models (MM-TSFM) for robustness (Section \ref{sec:rating}). 
2. Extend perturbation techniques used in other contexts to MM-TSFM, demonstrating their impact in three different settings and evaluating their robustness using our rating method across these settings: Input-specific Perturbation (IP), Semantic Perturbation (SP), and Compositional Perturbation (CP) (Section \ref{sec:perturbations}).
3. Demonstrate how stakeholders can utilize our ratings to choose a resilient MM-TSFM from the available alternatives, considering the data provided (see Table \ref{tab:cases}).
4. Find that our rating system suggests that semantic perturbations are the most disruptive compared to other perturbations for all the test systems (Section \ref{sec:expts}). 
5. Within the scope of our study, we conclude that multi-modal forecasting, which combines numeric and visual data and has been previously shown to outperform purely numeric forecasting, exhibit greater robustness as well (Table \ref{tab:ratings-pie}) with MM-TSFM showing 30 \% lower confounding bias and 60 \% lower MASE under semantic perturbations, compared to numerical forecasting.

In the remainder of the paper, we begin with a background on time-series forecasting and rating of AI systems, followed by the problem formulation and an introduction to our setup, which includes datasets, test systems, and perturbations. We present various hypotheses and support them with experimental results. We conclude with a discussion. The supplementary material contains rating algorithms adapted from \cite{kausik2024rating}, and some additional results.
\section{Related Work}
We now contextualize our work with related literature so that our contributions are highlighted. We cover MM-TSFM, adversarial attacks in finance, causal analysis in time-series forecasting, and rating of AI systems. 

\subsection{Multi-modal Time-series Forecasting}

Time-series forecasting is the problem of predicting a quantity at a future time given a history of the quantity indexed with time. 
This is a well-studied problem - we review only some recent advances.

In \cite{jia2024gpt4mts}, the authors propose a prompt-based LLM framework called GPT4MTS to utilize the numerical data and articles containing information about events. \cite{jin2023time} employs a reprogramming strategy to adapt Large Language Models (LLMs) for general time series forecasting without altering the underlying language model. \cite{ekambaram2020attention} addresses the challenge of forecasting sales for newly launched products using attention-based multi-modal encoder-decoder models that use product images and other provided attributes. \cite{cheng2022financial} introduces a multi-modal graph neural network to learn from multi-modal inputs. Additionally, \cite{yu2023temporal} leveraged multi-modal data including publicly available historical stock prices, company metadata, and historical economic news in their use of LLMs for conducting explainable time series forecasting. However, in these works, the data was collected from multiple sources (numerical, news articles, etc.) to train their models. In contrast, in \cite{zeng2023from}, the authors introduce a vision transformer that uses time-frequency spectrograms as the visual representation of time-series data transforming numerical data into a multi-modal form. Their experiments demonstrate the benefit of using spectrograms augmented with numeric time series and vision transformers for learning in both time and frequency domains compared to other existing image-based and numerical approaches. We compliment their work by proposing a method to rate different MM-TSFM for robustness using a causally grounded setup. Specifically, we evaluate two variations of ViT-num-spec model introduced in \cite{zeng2023from}.

\subsection{Adversarial Attacks in Finance Domain}
Malicious agents are incentivized to attack ML models deployed by financial institutions for financial gains, especially considering that the most efficient models utilize deep-learning architectures resembling those in the NLP community, which pose a challenge due to their vast number of parameters and limited robustness.

\cite{nehemya2021taking} shows the vulnerability of algorithmic trading systems to adversarial attacks by demonstrating how attackers can manipulate input data streams in real-time using imperceptible perturbations, 
highlighting the need for mitigation strategies in the finance community.
\cite{fursov2021adversarial} studies adversarial attacks on deep-learning models on transaction records, common in finance, demonstrating vulnerabilities. 
\cite{xu2021adversarial} was one of the first works to consider the vulnerability of deep learning models to adversarial time series examples in which they add noise to time-series plots to attack a ResNet model. 
However, none of these works evaluate multi-modal models against such attacks, nor do they explore multi-modal attacks or perturbations. In our work, we address several perturbations most of which can be applied to uni-modal or multi-modal systems.
We use both common data errors and attacks as motivations for the introduced input perturbations that we use to measure the robustness of time-forecasting models.


\subsection{Robustness Testing of Time-series Forecasting Models}
\cite{gallagher2022investigating} examines the impact of different attacks 
on the performance of Convolutional Neural Network model used for time series classification. \cite{pialla2023time} introduces a stealthier attack using the Smooth Gradient Method (SGM) for time series and measures the effectiveness of the attack.
While \cite{pialla2023time} focuses on measuring the smoothness of the attacks, our work quantifies their impact on the models in addition to the biases they create in models' predictions. \cite{govindarajulu2023targeted} adapt attacks from the computer vision domain to create targeted adversarial attacks. They examine the impact of the proposed targeted attacks versus untargeted attacks using statistical measures. 
All these works measure the models' performances under perturbations using statistical methods but do not measure the isolated impact of perturbations which is only possible through causal analysis. Furthermore, they do not consider any transformer-based or multi-modal models for evaluation.

\subsection{Causal Analysis in Time-series Forecasting}
\cite{moraffah2021causal} provides a review of the approaches used to compute treatment effects and also discusses causal discovery methods along with commonly used evaluation metrics and datasets. In this paper, they classify the task of estimating the effect of a treatment into three types: (a) Time-varying, (b) Time-invariant, and (c) Dynamic regimes. As the perturbations we apply to our data do not occur at the same timestep in each sample, the effect we are measuring can be considered as time-varying perturbation effect which is more complex to measure compared to time-invariant treatment effects. \cite{robins1999estimation} measures the causal effect of one such time-varying exposure. However, they only consider binary outcomes. In our work, we deal with continuous outcome attribute. \cite{bica2020time} proposes a time series deconfounder which deals with multiple treatment assignments over time and also takes into account hidden confounders. In contrast, we deal with one treatment at a time and focus on analyzing the treatment effects with respect to confounders that are of interest in the time-forecasting domain. 


\subsection{Rating AI Systems}
In the past, there were several works on assessing and rating different AI systems for trustworthiness from a third-party perspective (without access to the system's training data). In \cite{srivastava2018towards,srivastava2020rating,srivastava2023advances-rating}, the authors propose an approach to rate automated machine language translators for gender bias. Further, they create visualizations to communicate ratings \cite{bernagozzi2021vega}, and conduct user studies to determine how users perceive trust \cite{bernagozzi2021vega}. Though they were effective, they did not provide any causal interpretation. In another series of works \cite{kausik2022why, kausik2024rating, kausik2023the}, the authors introduced an approach to rate sentiment analysis systems for bias through causal analysis. 
We extend their method to rate MM-TSFM for robustness against perturbations. There are several advantages of using causal analysis over statistical analysis: causal models allow us to determine accountability and they align well with humanistic values.
Furthermore, they help us quantify the direct influence of various attributes on forecasting accuracy, isolating specific impacts, and mitigating confounding effects.


\section{Problem}
\label{sec:problem}
\subsection{Preliminaries}
\subsubsection{Stock Price Prediction:}
Let the time-series be represented by \{x$_{t-n+1}$, x$_{t-n+2}$, ...., x$_{t}$, x$_{t+1}$, ..., x$_{t+d}$\}, where each x$_{t-n+i}$ represents an adjusted closing stock price and $n$ and $d$ are  parameters. $n$ is called the sliding window size and $d$ is the number of future stock prices the model predicts.    
    In this sequence, let 
    X$_{t}$ $=$ \{x$_{t-n+1}$, x$_{t-n+2}$, ...., x$_{t}$\}, and let  $\hat{Y_{t}}$ $=$ \{x$_{t+1}$, 
    x$_{t+2}$, ...., x$_{t+d}$\}, where $\hat{Y_{t}}$ = f(X$_{t}$) i.e., based on the stock prices at previous $n$ timesteps, let the function $f$ represent an MM-TSFM that predicts the stock prices for the next `d' timesteps. Let $Y_{t}$ denote the true stock prices for the next `d' timesteps. Let $S$ be the set of time-series forecasting systems we want to assess. Let R$_{t}$ be the residual for the sliding window [t + 1, t + d] and is computed by ($\hat{Y_{t}}$ - $Y_{t}$) at each timestep. 

As the objective of our rating method is to consider the worst possible case (in this case, residual) to communicate the worst possible behavior of the model to the end-user, let us consider the MAX(R$_{t}$). Let it be denoted by $R^{max}_{t}$. 

\begin{figure}[!h]
    \centering
    \includegraphics[width=0.25\textwidth]{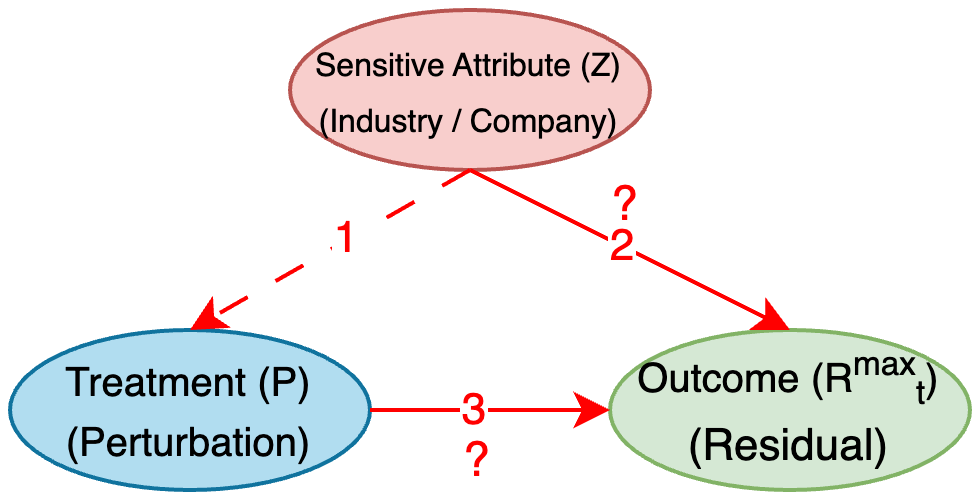}
	\caption{Causal diagram $\mathcal{M}$ for MM-TSFM. The validity of link `1' depends on the data distribution, while the validity of links `2' and `3' (with `?') are tested in our experiments.}
	\label{fig:causal-model}
\end{figure}

\vspace{-1em}
\subsubsection{Causal Analysis:}
Let the causal model that represents the system be denoted by \( \mathcal{M} \) - it is diagrammatically represented in  Figure \ref{fig:causal-model}. The arrowhead direction shows the causal direction from cause to effect. In Figure \ref{fig:causal-model}, if \emph{Sensitive Attribute} ($Z$) acts as a common cause for both \emph{Perturbation} ($P$) and \emph{Residual} ($R^{max}_{t}$), that would add a spurious correlation between the $P$ and the $R^{max}_{t}$. This is also called the confounding effect and \emph{Sensitive Attribute} is considered the confounder in this case. In the presence of a confounder, the path from \emph{Perturbation} to \emph{Residual} through the confounder is called a backdoor path and such a path is considered undesirable. Several backdoor adjustment techniques were used in the literature to remove the confounding effect \cite{xu2022neural, fang2024backdoor, liu2021preferences}. The deconfounded distribution (after backdoor adjustment) is represented by: $(R^{max}_{t} | do(P))$. The `do()' in causal inference denotes an intervention used to measure the causal effect of $P$ on the outcome. The solid red arrows in Figure \ref{fig:causal-model} denote the causal links whose validity and strength will be tested through our experiments. The dotted red arrow denotes the causal link that may or may not be present based on the distribution $(Perturbation | Sensitive Attribute)$ across different values of the \textit{Sensitive Attribute} i.e. if any perturbation that is applied depends on the values of $Z$. 

\subsection{Problem Formulation}
We aim to answer the following questions through our causal analysis when different perturbations denoted by $P$ = \{0, 1, 2, ..\} (or simply $P0, P1,...., P5$) are applied to the input given to the set of time-series forecasting systems $S$:

\noindent{\bf RQ1: Does $Z$ affect $R^{max}_{t}$ of $S$, even though $Z$ has no effect on $P$?}

\noindent That is, if the applied perturbations do not depend on the value of the sensitive attribute, would the sensitive attribute still affect the system outcome leading to a statistical bias? In this case, causal analysis is not required to answer the question as there is no confounding effect. 

\noindent {\bf RQ2: Does $Z$ affect the relationship between $P$ and $R^{max}_{t}$ of $S$ when $Z$ has an effect on $P$?}

\noindent That is, if the applied perturbations depend on the value of the sensitive attribute, would the sensitive attribute add a spurious (false) correlation between the perturbation and the outcome of a system leading to confounding bias? 

\noindent {\bf RQ3: Does $P$ affect $R^{max}_{t}$ of $S$ when $Z$ has an effect on $P$?}

\noindent That is, if the applied perturbations depend on the value of the sensitive attribute, what is the impact of the perturbation on the outcome of a system?

\noindent {\bf RQ4: Does $P$ affect $\hat{Y_{t}}$ of $S$ in the presence of $P$?}

\noindent That is, do the perturbations affect the performance of the systems in terms of accuracy? Causal analysis is not required to answer this question as we only need to compute appropriate accuracy metrics to assess how robust a system is against different perturbations.

\section{Method}
In this section, we discuss the architecture and working of MM-TSFM introduced in \cite{zeng2023from}. We then describe the different types of perturbations we used to assess MM-TSFM. We also describe the rating method tailored to our setting, originally introduced in \cite{kausik2024rating}. 

\subsection{Multi-modal Time-series Forecasting Models (MM-TSFM)}
\label{sec:models}
In this study, we employ the $\textbf{ViT-num-spec}$ model \cite{zeng2023from}, which combines a \textbf{\underline{vi}}sion \textbf{\underline{t}}ransformer with a multimodal time-frequency \textbf{\underline{spec}}trogram, augmented by the intensities of \textbf{\underline{num}}eric time series for time series forecasting. This model enhances predictive accuracy by leveraging both visual and numerical data. Specifically, it transforms numeric time series into images using a time-frequency spectrogram and utilizes a vision transformer (ViT) encoder with a multilayer perceptron (MLP) head for future predictions. We trained two variations of the ViT-num-spec model using two real-world datasets and conducted evaluations using a separate dataset.


\subsubsection{Time-Frequency Spectrogram} \label{subsec:spectrogram}
Building on the method of \cite{zeng2023from}, we use wavelet transforms \cite{daubechies1990wavelet} to create time-frequency spectrograms from time series data. 
Specifically, we employ the Morlet wavelet (see \cite{morlet2}) with scale $s$, and $\omega_0 = 5$ by default, as detailed in Equation \ref{wavelet_eq}. The parameter $s$ represents the scale of the wavelet, affecting its width and localization in time, while $\omega_0$ is the central frequency of the wavelet, influencing its oscillatory behavior. The wavelet function $\psi(x)$ is given by:
{
\tiny
\begin{equation}
\label{wavelet_eq}
\psi(x) = \sqrt{\frac{1}{s}} \pi^{-\frac{1}{4}} \exp\left(-\frac{x^2}{2s^2}\right) \exp\left(j\omega_0\frac{x}{s}\right)
\end{equation}
}

This method involves convolving the time series with wavelets at various scales. The magnitudes of the resulting coefficients indicate the strength of the signal at different frequencies. These magnitudes are visualized in a spectrogram, with higher frequencies positioned at the top and lower ones at the bottom. To enhance the utility of these spectrograms, an image stripe is added at the top of the spectrogram image to retain the sign information of the time series. Specifically, this image stripe is converted from the standardized numeric time series, into a stripe of pixels with intensity values ranging from 0 to 255. This enhanced image is then used as the input for the vision transformer model, as discussed next.

\subsubsection{Vision Transformer}\label{subsec:visiontransformer}
In the next stage, the ViT-num-spec employs a vision transformer architecture that integrates a MLP head for the purpose of forecasting time series. The input images are segmented into non-overlapping patches of uniform size, effectively partitioning the horizontal time axis into discrete intervals corresponding to each image patch. These patches undergo linear projection to be transformed into tokens, augmented with standard 1D positional embeddings to maintain temporal ordering. The encoder is tasked with converting these patches into latent representations or feature vectors. For our implementation, we configure the input image dimensions to 128×128 pixels, with each patch sized at 16×16 pixels. The time series data from price movements is formatted to fit a 16×128 top row, while the spectrogram occupies a 112×128 space below it, ensuring compatibility with the patch size for optimal processing.

\begin{figure*}[!ht]
    \centering
    \includegraphics[width=\textwidth, height=11em]{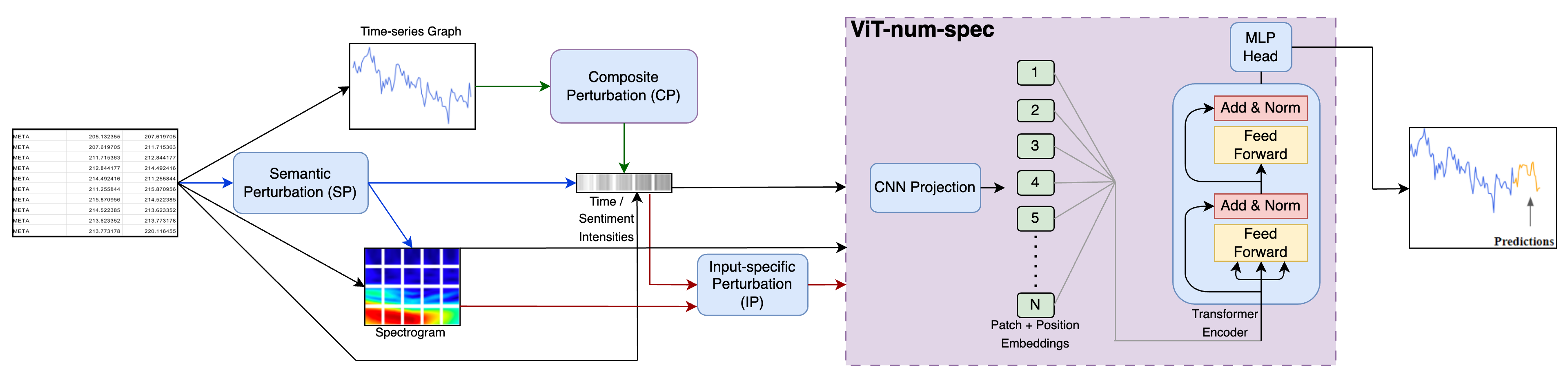}
    \caption{
    Overview of our proposed `data to predictions' workflow with ViT-num-spec \cite{zeng2023from} showing different perturbation pathways: black arrows for no perturbations, blue for semantic (SP), green for composite (CP), and red for input-specific (IP). Predictions from these pathways are rated in the next step.
    }
	\label{fig:system-workflow}
\end{figure*}

\begin{figure*}[!ht]
    \centering
    \includegraphics[width=\textwidth, height=12em]{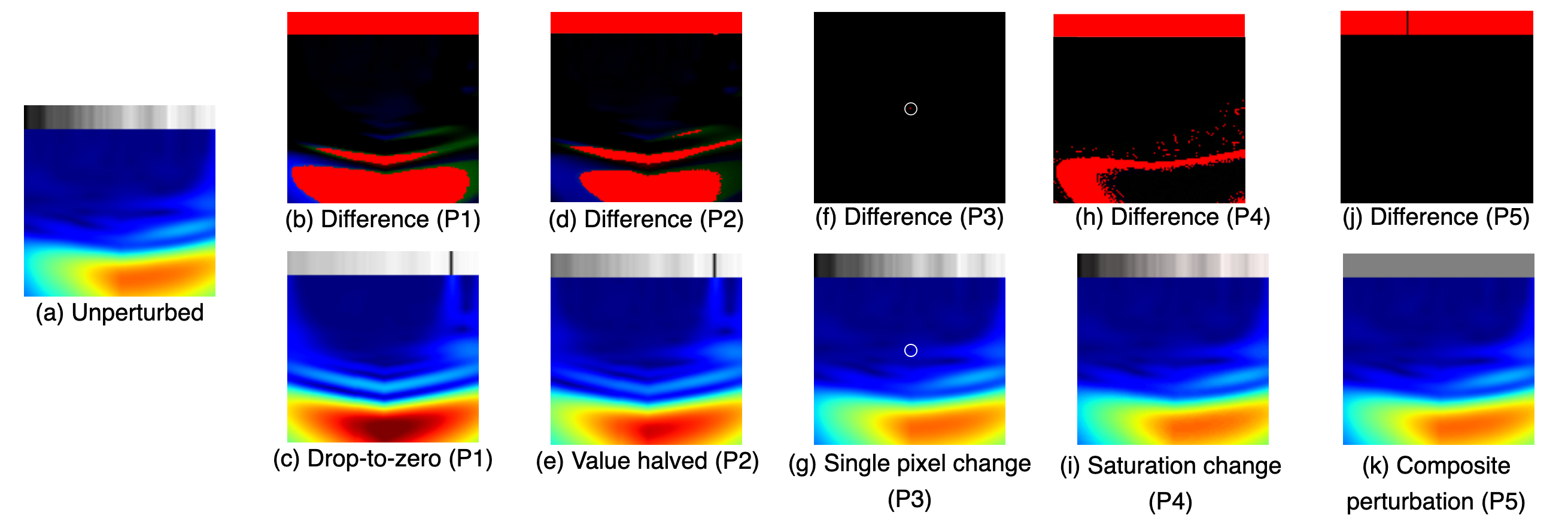}
    \caption{Unperturbed image and its perturbed variations, along with a corresponding difference image that highlights (in red) the portion that is modified after perturbation. In (f,g), the single pixel change is highlighted with a white circle around it.
    }
    \label{fig:perturbations}
    \end{figure*}

\begin{figure}[!ht]
    \centering
    \includegraphics[width=0.35\textwidth, height=13em]{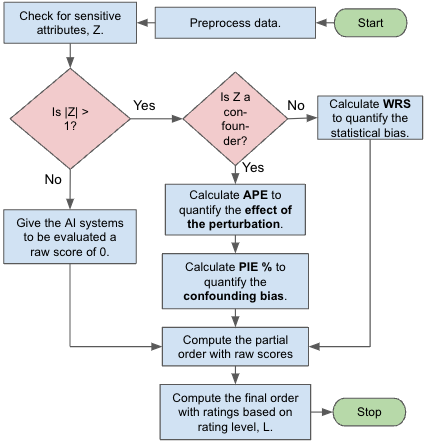}
    \caption{Our proposed 'predictions to ratings' workflow that performs statistical and causal analysis to compute raw scores and assign final ratings to the test systems.}
	\label{fig:rating-workflow}
\end{figure}

\subsection{Perturbations}
\label{sec:perturbations}
We assess the robustness of the models described in \nameref{sec:models} section by subjecting the input data to different perturbations across three different settings:
\subsubsection{Semantic Perturbation (SP)}
Semantic perturbations are alterations made to data that change its meaning while preserving the overall context.
For example, in time-series forecasting, a stock's value might change drastically due to errors in data entry, or it might fluctuate due to some market-specific catalyst that affects certain companies. Under SP, we consider two perturbations:

\noindent {\bf 1. Drop-to-zero (P1):} It is inspired by 
the common data entry errors \cite{ley2019analysis}. We set every $80^{th}$ value in the original stock price data collected from the source to zero. Once converted into a sliding window format with 80 timesteps per row, this approach ensures that each row contains at least one zero.

\noindent {\bf 2. Value halved (P2):} In this setting, we reduce every $80^{th}$ number in the original stock price data to half of its value. This perturbation simulates periodic adjustments, possibly reflecting events like stock splits or dividend payments.

\subsubsection{Input-specific Perturbation (IP)}
Input-specific perturbations are alterations specific to the mode - features and context of the data being used. In time-series forecasting, altering some pixels (e.g., changing their color) in a time-frequency spectrogram image is one example. Under IP, we consider two different perturbations:

\noindent {\bf 1. Single pixel change (P3):} In \cite{su2019one}, the authors modified a single pixel in each of the test images. With this approach, they fooled three types of DNNs trained on the CIFAR dataset. 
In our work, we alter the center pixel of each multi-modal input to black based on the intuition that 
small and consistent change to the images can significantly alter the models' prediction.

\noindent {\bf 2. Saturation change (P4):} In \cite{zhu2023imperceptible}, the authors showed that the adversarial perturbations in the S-channel (or saturation channel) of an image in HSV (Hue Saturation Value) form ensures a high success rate for attacks compared to other channels. 
In our work, we increase the saturation of the multi-modal input ten-fold 
based on the intuition that a subtle change can affect the models' predictions.

\subsubsection{Composite Perturbation (CP or P5)}
We consider a composite case in which the MM-TSFM is combined with another AI system. This is because it is well known in finance that stock prediction can be influenced by market sentiment \cite{senti-bias-finance} and we want to explore such a realistic case. In \cite{kausik2023the}, the authors considered a composite case in which the input text is round-trip translated (translation from one language to the same through another intermediate language) and passed to a sentiment analysis system (SAS). Then they used the rating method to assess the bias in the system. In our work, we assess the sentiment of each time series by passing the corresponding time-series plot to a zero-shot CLIP-based \cite{radford2021learning, bondielli2021leveraging} sentiment analysis system ($S_c$) 
which outputs the sentiment intensity values (negative (-1), neutral (0), or positive (1)). These labels are scaled to values in the range [0, 255] and represented as sentiment intensity stripe and passed to MM-TSFM in place of the original time-series intensity stripe. This zero-shot CLIP-based $S_c$ for sentiment analysis may exhibit bias or may be inaccurate, the goal is to study the robustness of MM-TSFM in conjunction with such AI system.

\subsection{Workflow}
\label{sec:rating}

\subsubsection{From Data to Predictions:}

Figure \ref{fig:system-workflow} shows the overview of the proposed workflow with the ViT-num-spec model as the MM-TSFM. The paths highlighted in different colors correspond to different perturbations. The semantic perturbations (highlighted in blue), when applied to the numerical time-series data, are reflected in the corresponding spectrogram and the time-intensity images. The composite perturbation (highlighted in green), when applied, gets reflected in the corresponding time-series intensity image stripe. The spectrogram remains unperturbed. The input-specific perturbation (highlighted in red), directly affects both the spectrogram and the time intensities images. These perturbed/unperturbed input data points are fed to the MM-TSFM. The model processes the input (as described in \nameref{sec:models}) and predicts the next `d' timesteps (we set $d = 20$).

\subsubsection{From Predictions to Ratings:}
In \cite{kausik2024rating}, the authors introduced an approach to rate text-based sentiment analysis systems for bias. The rating algorithms we adapted from their paper are provided in Section 2 of the supplementary. We modify this approach to better suit our setting as it involves multi-modal data which is more complex than the textual data that was used in the original rating work. Also, our approaches involve multiple treatments (perturbations). Figure \ref{fig:rating-workflow} shows the overview of the proposed rating workflow. After preprocessing the data (optional), the first step is to check for sensitive attributes that may act as confounders. If they do not act as confounders, statistical methods to answer RQ1 from Section \nameref{sec:problem} would suffice. The metric used to assess the statistical bias is described in Section \ref{sec:metrics}. If they do act as confounders, causal analysis is necessary. We use two metrics: one to assess the impact of each perturbation and another to compute the confounding bias (to answer RQ2 and RQ3) respectively. These metrics are collectively referred to as raw scores (that form a partial order together) which aid in assigning the final ratings to the systems (that form a final order together). The range of final ratings depend on the value of rating level, L. The final ratings computed from our experiments are shown in Table \ref{tab:ratings-pie}.

\section{Experiments and Analysis}
In this section, we describe the experimental apparatus that was used to perform the experiments. It includes the description of test systems, evaluation/test data, and evaluation metrics. We also state and prove certain hypotheses that help us answer the research questions stated in Section \ref{sec:problem}.
\subsection{Experimental Apparatus}
\subsubsection{Test Systems Considered}
We evaluated the following time-series forecasting systems for robustness:

\noindent (a) \textbf{ViT-num-spec}($S_v$): We trained two models on the daily stock time series of S\&P 500 constituents: one using data from before the COVID-19 outbreak and another with data from during the pandemic. We assessed our methodologies by comparing predictions from these models, each trained on datasets from distinct time periods.

\noindent(i) \textbf{ViT-num-spec trained on pre-COVID data  ($S_{v1}$):} The pre-COVID model was trained using 46,875 time series and validated on a separate set of 46,857 time series, spanning from 2000 to 2014. The vision transformer encoder outputs a latent embedding vector with a dimensionality of 128. Training involved a batch size of 128 and employed the AdamW optimizer with a weight decay of 0.05. The process spanned up to 200 epochs, incorporating an early stopping mechanism set with a patience of 10 epochs. Additionally, the base learning rate and learning rate scheduler hyperparameters were finely tuned for the dataset to optimize performance.

\noindent(ii)\textbf{ViT-num-spec trained on during-COVID data ($S_{v2}$)}: 
The during-COVID model was trained on a dataset comprising 7,478 time series, and validated on a separate set of 7,475 time series, covering the period from March 2020 to November 2022. It shares the same architecture and parameter tuning routine as the model trained on pre-COVID data, utilizing 80 past time steps to predict the subsequent 20 steps.

\noindent(b) \textbf{ARIMA} ($S_a$): ARIMA (AutoRegressive Integrated Moving Average) model is a widely used statistical approach for time series forecasting. It combines three different components: Autoregressive (AR), differencing (I), and moving average (MA) to capture the patterns in the time-series data and predict the next `d' (from section \ref{sec:problem}) values. To make the time-series data stationary, we had to difference it once. Using ACF and PACF plots, we were able to identify the optimal p and q values. 

\noindent(c) \textbf{Biased system} ($S_b$): We built this system as an extreme baseline. $S_b$ is more biased towards META and GOOG (technology companies) compared to other companies. The system assigns a residual of `0' and `200' for META and GOOG, respectively, while assigning higher residuals to other companies. This ensures that the system shows maximum amount of bias in each setting we considered, representing an extreme baseline for evaluating the performance of other systems. 

\noindent(d) \textbf{Random system} ($S_r$): We built this system as another extreme baseline. $S_r$ assigns random price predictions within the range [MIN(company's stock prices) - 100, MAX(company's stock prices) + 100] based on each company's stock prices. Although the values generated depend on the range of the stock prices for each company, considering the residuals as the outcome, will still add sufficient randomness to the generated numbers. It will also ensure that the random values are contextually meaningful.

\subsubsection{Evaluation/Test Data}
We collected daily stock prices from Yahoo! Finance for six companies: Meta Platforms, Inc. (META), Google (GOO), Pfizer Inc. (PFE), Merck (MRK), Wells Fargo (WFC), and Citigroup Inc. (C). 
These companies belong to different sectors with META, GOO in social technology, PFE, MRK in pharmaceuticals, and WFC, C in financial services. The data spans from March 28, 2023, to April 22, 2024. We used a subset of this data, from March 28, 2023, to March 22, 2024, to predict stock prices for the following month. These companies, with their high stock prices, are representative of their respective industry. We applied the sliding window ($width=1$) technique to sample (input, output) pairs for the prediction task. Specifically, the input time series contains 80-days stock prices, and the output contains 20-days stock prices (here, n=80 and d=20 (from section \ref{sec:problem})). 


\subsection{Evaluation Metrics}
\label{sec:metrics}
In this section, we define our evaluation metrics. We adapt WRS metric originally proposed in \cite{kausik2024rating}. Additionally, we introduce two new metrics: APE and PIE \% (modified versions of ATE \cite{abdia2017propensity} and DIE \% \cite{kausik2024rating}) tailored to answering RQ2 and RQ3.

\noindent{\bf Weighted Rejection Score (WRS):} WRS was introduced in \cite{kausik2024rating} as a way to measure statistical bias. First, Student's t-test \cite{student1908probable} is used to compare different max residual distributions $(R^{max}_{t} | Z)$ for different values of protected attribute, $Z$. In our case, we measure this between each pair of industries or companies resulting in $^3C_2 = 3$ comparisons. 
Next, the computed t-value of each pair of companies is compared with the critical t-value 
(computed based on confidence interval (CI) and degrees of freedom (DoF)). Based on this, the null hypothesis is either rejected or accepted. 

    The t-value is computed using the following equation:
    $$t = (mean(x_1) - mean(x_2))/\sqrt{((s_1^2/n_1)+(s_2^2/n_2))}$$

    where $x_1$ and $x_2$ are two $(R^{max}_{t} | Z)$ distributions, $s_1$ and $s_2$ are their standard deviations, and $n_1$, $n_2$ are the sample sizes. 
    \cite{kausik2024rating} added a small $\epsilon$ value to the denominator in cases to handle cases where standard deviation becomes zero. They chose different confidence intervals (CI) [95\%, 75\%, 60\%] that have different critical values and if the computed t-value is less than the critical value, the null hypothesis was rejected. WRS is mathematically defined by the following equation:


\begin{equation}
\textbf{Weighted Rejection Score (WRS)} = \sum_{i\in CI} w_i*x_i,
\label{eq:wrs}
\end{equation}
    \noindent where, $x_i$ is the variable set based on whether the null hypothesis is accepted (0) or rejected (1). $w_i$ is the weight that is multiplied by $x_i$ based on the CI. For example, if CI is 95\%, $x_1$ is multiplied by 1. The lower the CI, the lower the weight will be. WRS helps us answer RQ1 (see Section~\ref{sec:problem}). 
    \vspace{0.2cm}
    



\noindent{\bf Average Perturbation Effect (APE)}: Average Treatment Effect (ATE) provides the average difference in outcomes between units in the treatment and units in the control \cite{wang2017g}. In our context, it computes the difference between residuals between perturbed data points (P1 through P5) and the data points in the control (P0). Hence, we refer to this metric as APE. This metric helps us answer RQ3. It is formally defined using the following equation:

    {\bf APE} = 
    {\small
        \begin{equation}
        \begin{split}
        [|E[R^{max}_{t} = j| do(P = i)] - E[R^{max}_{t} = j| do(P = 0)]| ]
        \label{eq:ape}
        \end{split}
        \end{equation}
    }

    
\noindent {\bf Propensity Score Matching - Deconfounding Impact Estimation \% (PSM-DIE \% or in short, PIE \%):} In \cite{kausik2024rating}, the authors used a linear regression model to estimate the causal effects. This model assumes that the relationship between the treatment variables, covariates, and outcome is linear. The linear regression method does not capture non-linear relationships and may not fully eliminate confounding biases. They proposed a metric called DIE \% that computes relative difference between deconfounded and confounded distributions using binary treatment values. However, we use six different treatment (or perturbation) values in our work. We use the Propensity Score Matching (PSM) \cite{rosenbaum1983central} technique that specifically targets the confounding effect by matching the treatment and control units based on the probability of receiving a treatment, making both groups comparable. PSM is analogous to randomized controlled trials (RCTs) and does not involve the outcome variables \cite{baser2007choosing}, whereas linear regression does. We modify DIE \% and introduce the PIE \% (PSM-DIE\%) metric. This metric helps us answer RQ2. It is given by the formula:

    {\small
        \begin{equation}
        \textbf{PIE \%} = \left[ ||APE_{o}| - |APE_{m}|| \right] * 100
        \label{eq:pie}
        \end{equation}
    }
    
   $APE_{o}$ and $APE_{m}$ represent the APE computed before and after applying PSM respectively. $PIE \%$ computes the true impact $Z$ has on the relation between $P$ and $R^{max}_{t}$.


\noindent {\bf Symmetric mean absolute percentage error (SMAPE) } is defined as, 
    {\small
    \begin{equation} \label{eq:smape}
        SMAPE = \frac{1}{T}\sum_{t=1}^T\frac{|x_t - \hat{x}_t|}{(|x_t| +|\hat{x}_t|)/2}, 
    \end{equation}
    }
where $T= 20$ is the total number of observations in the predicted time series, $x_t$ represents the actual observed values, and $\hat{x}_t$ the predicted values at each time step $t= 1, \ldots, 20$. SMAPE scores range from 0 to 2, with lower scores indicating more precise forecasts. 

\noindent {\bf Mean absolute scaled error (MASE)} quantifies the mean absolute error of the forecasts relative to the mean absolute error of a naive one-step forecast calculated on the training data.
{\small
\begin{equation}\label{eq:mase}
    MASE=\frac{\frac{1}{T} \sum_{i=t+1}^{t+T}|x_{i} - \hat{x_{i}}|}{\frac{1}{t}\sum_{i=1}^{t}|x_{i} - x_{i-1}|},
\end{equation}
}
where in our case, $t = 80$, and $T = 100$.
Lower MASE values indicate better forecasts. 

\noindent {\bf Sign Accuracy} quantifies the average classification accuracy across all test samples, where a higher accuracy indicates more precise predictions. This metric classifies based on how the predicted forecasts align with the most recent observed values in the input time series.

\subsection{Experimental Evaluation}
\label{sec:expts}
In this section, we state hypotheses, the experimental setup used, the results obtained, interpretations made from the results, and the conclusions drawn from the interpretations. Table \ref{tab:cases} summarizes the research questions answered through the hypotheses stated along with the corresponding causal diagrams and key findings from these hypotheses. Some of the experimental results can be found in the supplementary. Table 3 and 4 in the supplementary show partial order and final order with respect to WRS, APE, SMAPE, and Sign Accuracy.

\begin{table*}[ht]
\centering
   {\tiny
    \begin{tabular}{|p{14em}|p{12em}|p{14em}|p{4em}|p{30em}|}
    \hline
          {\bf Research Question} &    
          {\bf Hypothesis} & 
          {\bf Causal Diagram} &
          {\bf Metrics Used} &
          {\bf Findings} \\ \hline 
          
          \textbf{RQ1:} Does $Z$ affect $R^{max}_{t}$ of $S$, even though $Z$ has no effect on $P$? & 
          \emph{Company} affects the \emph{Residual} of \emph{S}, even though \emph{Company} has no effect on \emph{Perturbation}. &
          \begin{minipage}{.05\textwidth}
          \vspace{2.5mm}
          \centering
          \includegraphics[width=30mm, height=15mm]{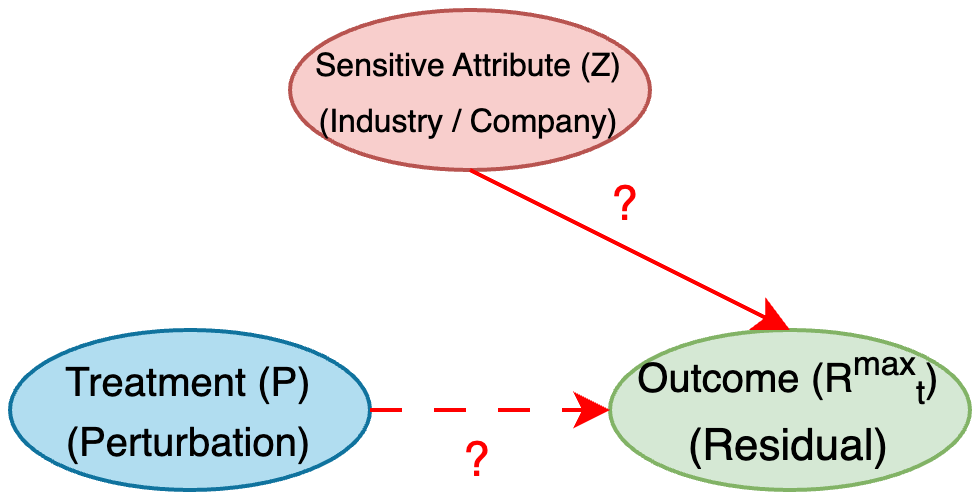}
          \end{minipage} &
          WRS &
          (Table 2 and 3 in the supplementary) \textbf{\textit{S} with low statistical bias}: $S_a$ and $S_{v2}$. \textbf{\textit{P} that led to more statistical bias}: P1 and P2.  \textbf{Analysis with more discrepancy}: Inter-industry
          \\ \hline 
          \textbf{RQ2:} Does $Z$ affect the relationship between $P$ and $R^{max}_{t}$ of $S$ when $Z$ has an effect on $P$?& 
          \emph{Company} affects the relationship between \emph{Perturbation} and \emph{Residual} of \emph{S} when \emph{Company} has an effect on \emph{Perturbation}. &
          \begin{minipage}{.05\textwidth}
          \vspace{2.5mm}
          \centering
          \includegraphics[width=30mm, height=15mm]{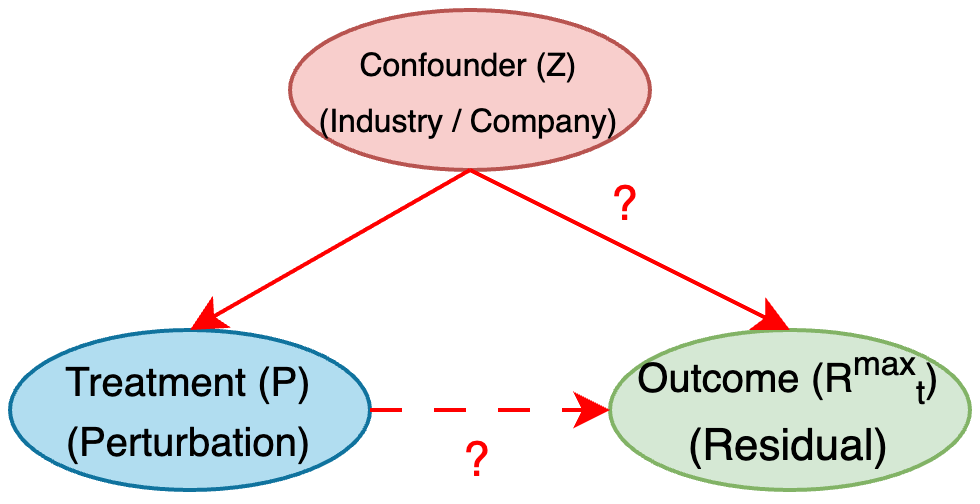}
          \end{minipage} &
          PIE \% &
          (Table 2 in supplementary and Table \ref{tab:ratings-pie}) \textbf{\textit{S} with low confounding bias}: $S_{v2}$ and $S_{v1}$. \textbf{\textit{P} that led to more confounding bias}: P1 and P2.  \textbf{Confounder that led to more bias}: \textit{Industry}
          \\ \hline 
          \textbf{RQ3:} Does $P$ affect $R^{max}_{t}$ of $S$ when $Z$ has an effect on $P$? & 
          \emph{Perturbation} affects the \emph{Residual} of \emph{S} when \emph{Company} has an effect on \emph{Perturbation}. &
          \begin{minipage}{.05\textwidth}
          \vspace{2.5mm}
          \centering
          \includegraphics[width=30mm, height=15mm]{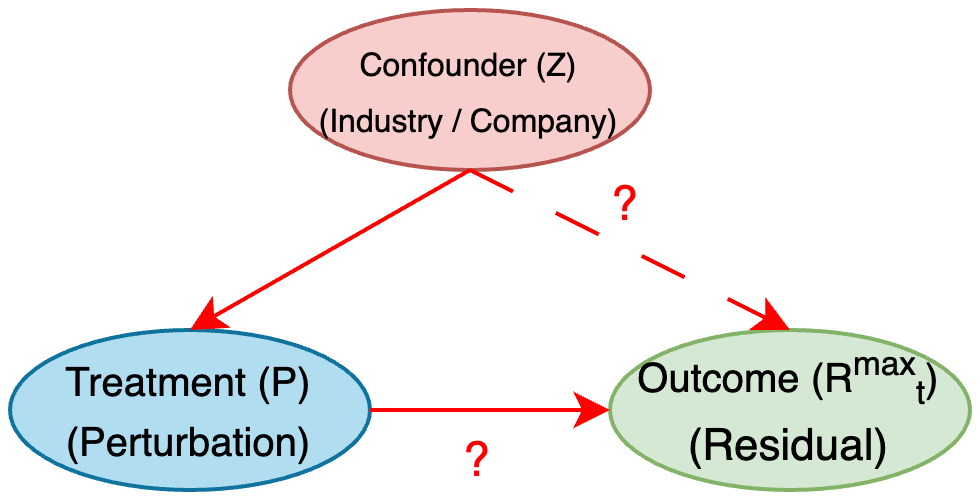}
          \end{minipage} &
          APE &
          (Table 2 and 4 in the supplementary) \textbf{\textit{S} with low APE}: $S_{v1}$. \textbf{\textit{P} with high APE}: P1 and P2.  \textbf{Confounder that led to high APE}: \textit{Industry}
          \\ \hline 
          \textbf{RQ4:} Does $P$ affect $\hat{Y_{t}}$ of $S$ in the presence of $P$? & 
          Semantic and compositional \emph{Perturbations} can degrade the performance of \emph{S}.  &
          This hypothesis does not necessitate a causal model for its evaluation. &
          SMAPE, MASE, Sign Accuracy &
          (Table \ref{tab:h4}) \textbf{\textit{S} with good performance}: $S_{v1}$. \textbf{\textit{P} with high impact on performance}: P1, P2 and P5.  
          \\ \hline 
    \end{tabular}
    }
    \caption{Summary of the research questions answered in the paper along with corresponding hypotheses stated, causal diagram, metrics used in the experiment, and the important findings from the experiment.
    }
    \label{tab:cases}
\end{table*}

\begin{table}[!ht]
\centering
{\tiny
    \begin{tabular}{|p{4.5em}|p{0.5em}|p{15em}|p{12em}|}
    \hline
          {\bf Forecasting Evaluation Dimensions} &
          {\bf P} &    
          {\bf Partial Order} &
          {\bf Complete Order} 
          \\ \hline 
          \multirow{5}{6em}{Confounding Bias with \textit{Industry} as confounder (PIE$_I$ \%)} &
          P1 & 
          \{$S_{v1}$: 630.10, $S_a$: 982.38, $S_{v2}$: 1191.91, $S_r$: 4756.40,  $S_b$: 6916.11\} &
          \{\textbf{S$_{v1}$: 1}, $S_a$: 1, $S_{v2}$: 2, $S_r$: 2,  $S_b$: 3\} 
          \\ \cline{2-4}
          &
          P2 & 
          \{$S_{v1}$: 941.93, $S_a$: 1275.04, $S_{v2}$: 1490.65, $S_r$: 4274.38,  $S_b$: 9474.61\} &
          \{\textbf{S$_{v1}$: 1}, $S_a$: 1, $S_{v2}$: 2, $S_r$: 2,  $S_b$: 3\} 
          \\ \cline{2-4}
          &
          P3 & 
          \{$S_{v2}$: 224.98, $S_{v1}$: 276.86, $S_r$: 3560.94,  $S_b$: 7489.48\} &
          \{\textbf{S$_{v2}$: 1}, $S_{v1}$: 1, $S_r$: 2,  $S_b$: 3\} 
          \\ \cline{2-4}
          &
          P4 & 
          \{$S_{v1}$: 229.03, $S_{v2}$: 1694.57, $S_r$: 2250.35,  $S_b$: 7618.25\} &
          \{\textbf{S$_{v1}$: 1}, $S_{v2}$: 1, $S_r$: 2,  $S_b$: 3\} 
          \\ \cline{2-4}
          &
          P5 & 
          \{$S_{v2}$: 273.12, $S_{v1}$: 344, $S_r$: 4025.31,  $S_b$: 8966.57\} &
          \{\textbf{S$_{v2}$: 1}, $S_{v1}$: 1, $S_r$: 2,  $S_b$: 3\} 
          \\ \hline
          \multirow{5}{6em}{Confounding Bias with \textit{Company} as confounder (PIE$_C$ \%)} &
          P1 & 
          \{$S_{v2}$: 415.74, $S_{v1}$: 551, $S_a$: 914.64, $S_r$: 1041.01,  $S_b$: 3283.88\} &
          \{\textbf{S$_{v2}$: 1}, $S_{v1}$: 1, $S_a$: 2, $S_r$: 2,  $S_b$: 3\} 
          \\ \cline{2-4}
          &
          P2 & 
          \{$S_{v2}$: 575.12, $S_{v1}$: 898.90, $S_a$: 1154.87, $S_r$: 1463.71,  $S_b$: 2174.39\} &
          \{\textbf{S$_{v2}$: 1}, $S_{v1}$: 1, $S_a$: 2, $S_r$: 2,  $S_b$: 3\} 
          \\\cline{2-4}
          &
          P3 & 
          \{$S_{v2}$: 1277.44, $S_r$: 1305.78,   $S_b$: 1846.56, $S_{v1}$: 2427.35\} &
          \{\textbf{S$_{v2}$: 1}, $S_r$: 1, $S_b$: 2,  $S_{v1}$: 3\} 
          \\ \cline{2-4}
          &
          P4 & 
          \{$S_{v1}$: 247.80, $S_{v2}$: 942.02, $S_r$: 1314.82,  $S_b$: 3557.45\} &
          \{\textbf{S$_{v1}$: 1}, $S_{v2}$: 1, $S_r$: 2,  $S_b$: 3\} 
          \\ \cline{2-4}
          &
          P5 & 
          \{$S_{v2}$: 284.95, $S_{v1}$: 378.19, $S_r$: 1928.21,  $S_b$: 2118.88\} &
          \{\textbf{S$_{v2}$: 1}, $S_{v1}$: 1, $S_r$: 2,  $S_b$: 3\} 
          \\\hline
          \multirow{5}{6em}{Accuracy (MASE)} &
          P0 & 
          \{$S_{v1}$: 3.68, $S_a$: 3.79, $S_{v2}$: 3.89, $S_r$: 86.45,  $S_b$: 947.56\} &
          \{\textbf{S$_{v1}$: 1}, $S_a$: 1, $S_{v2}$: 2, $S_r$: 2,  $S_b$: 3\} 
          \\ \cline{2-4}
          &
          P1 & 
          \{$S_{v1}$: 5.30, $S_{v2}$: 11.18, $S_a$: 18.36, $S_r$: 86.99,  $S_b$: 947.56\} &
          \{\textbf{S$_{v1}$: 1}, $S_{v2}$: 1, $S_a$: 2, $S_r$: 2,  $S_b$: 3\} 
          \\ \cline{2-4}
          &
          P2 & 
          \{$S_{v1}$: 4.24, $S_{v2}$: 6.16, $S_a$: 8.24, $S_r$: 86.87,  $S_b$: 947.56\} &
          \{\textbf{S$_{v1}$: 1}, $S_{v2}$: 1, $S_a$: 2, $S_r$: 2,  $S_b$: 3\} 
          \\ \cline{2-4}
          &
          P3 & 
          \{$S_{v1}$: 3.68, $S_{v2}$: 3.89, $S_r$: 86.65,  $S_b$: 947.56\} &
          \{\textbf{S$_{v1}$: 1}, $S_{v2}$: 1, $S_r$: 2,  $S_b$: 3\} 
          \\\cline{2-4}
          &
          P4 & 
          \{$S_{v1}$: 3.67, $S_{v2}$: 3.90, $S_r$: 86.53,  $S_b$: 947.56\} &
          \{\textbf{S$_{v1}$: 1}, $S_{v2}$: 1, $S_r$: 2,  $S_b$: 3\} 
          \\ \cline{2-4}
          &
          P5 & 
          \{$S_{v2}$: 3.93, $S_{v1}$: 8.26, $S_r$: 87.20,  $S_b$: 947.56\} &
          \{\textbf{S$_{v2}$: 1}, $S_{v1}$: 1, $S_r$: 2,  $S_b$: 3\} 
          \\ \hline
    \end{tabular}
    }
    \caption{Final raw scores and ratings based on PIE \% and MASE. Higher rating indicate higher bias (or error, for MASE). For simplicity, we denoted the raw scores for MASE using just the mean value, but standard deviation was also considered for rating. The chosen rating level, L = 3. From P1 and P2 results, it can be observed that for S$_{v1}$ the PIE$_I$ is almost 30 \% lower and MASE is 60 \% lower on average compared to $S_a$.}
    \label{tab:ratings-pie}
\end{table}

{\color{blue}\noindent{\bf Hypothesis-1:}} \emph{Sensitive Attribute} affects the \emph{Residual} of \emph{S}, even though \emph{Sensitive Attribute} has no effect on \emph{Perturbation}.

\noindent{\bf Experimental Setup:}
In this experiment, the causal link from the \emph{Sensitive Attribute} to \emph{Perturbation} will be absent as the application of perturbation to the stock prices does not depend on the corresponding company name or the industry i.e., perturbations are applied uniformly across all the data points. The corresponding causal diagram is shown in table \ref{tab:cases}. We quantify the statistical bias exhibited by the systems by using WRS that was defined in equation \ref{eq:wrs}. To compare \emph{Residual} distributions across different \emph{Sensitive Attribute} stock prices, we use t-value, p-value, and DoF from the Student's t-test \cite{student1908probable} to compare different outcome distributions across each of the companies. We perform two different analyses in this experiment: one to measure the discrepancy $S$ shown across different industries (WRS$_{Industry}$) and another to measure the discrepancy $S$ among both the companies (WRS$_{Company}$) within the same industry.

\noindent{\bf Results:} Table 1 in the supplementary shows the t-statistic computed for all distributions along with the number of rejections under each confidence interval and Table 2 in the supplementary shows the Weighted Rejection Scores (WRS) across companies within the same industry (WRS$_I$) and different industries (WRS$_C$) that quantify the statistical bias exhibited by different test systems. The final ratings with respect to the WRS are shown in Table 3 in the supplementary.

\noindent{\bf Interpretation:}  From Table 2 in the supplementary, most discrepancies can be observed across industries compared to the discrepancies across companies i.e., all the test systems make more prediction errors when compared across different industries. When the input data was subjected to perturbations P1 and P2, the systems exhibited more statistical bias. From Table 3 in the supplementary, $S_{v2}$ and $S_a$ exhibited the least amount of statistical bias under most of the perturbations compared to other systems. 

\noindent{\bf Conclusion:} $S_a$ and $S_{v2}$ outperformed other systems in terms of statistical bias. Systems exhibited more bias when P1 and P2 were applied and showed more statistical bias when compared across different industries rather than different companies within the same industry. Hence, \emph{Sensitive Attribute} affects the \emph{Residual} of \emph{S}, even though \emph{Sensitive Attribute} has no effect on \emph{Perturbation}. 

{\color{blue}\noindent{\bf Hypothesis-2:}} \emph{Confounder} affects the relationship between \emph{Perturbation} and \emph{Residual} of \emph{S} when \emph{Confounder} has an effect on \emph{Perturbation}.

\noindent{\bf Experimental Setup:} In this experiment, we use PIE \% defined in equation \ref{eq:pie} to compare the APE (defined in equation \ref{eq:ape}) before and after deconfounding using the Propensity Score Matching (PSM) technique as the presence of the confounder opens a backdoor path from \emph{Perturbation} to  \emph{Residual} through the \emph{Confounder}. The causal link from \emph{Confounder} to \emph{Perturbation} will be valid only if the perturbation applied depends on the value of the confounder (i.e. the company or the industry the specific data points belong to). To ensure the probability of perturbation assignment varies with respect to the \emph{Confounder} across three distributions (DI1 through DI3) in the case of \emph{Industry} and six different distributions in the case of \emph{Company} (DC1 through DC6), we implemented weighted sampling. For each distribution, weights were configured so that perturbation groups P1 through P5 had a twofold higher likelihood of selection compared to P0 for specific values of the confounder. For example, META in DC1, GOOG in DC2, and so on. This strategy highlights significant cases, although other combinations are possible for further exploration. Computing APE involves the following steps: First, we use logistic regression to estimate the probability of each data point receiving the perturbation based on the company or the industry it belongs to. Next, we match each perturbed data point with a control (not perturbed) data point that was not perturbed but has a similar probability of having perturbed. This matching process ensures that both groups are comparable, allowing us to isolate and measure the actual impact of the perturbation.

\noindent{\bf Results:}
Table 2 in the supplementary shows the PIE$_C$ \% (when \textit{Company} acts as the confounder) and PIE$_I$ \% (when \textit{Industry} acts as the confounder) values computed for different systems across different distributions. As our rating method aims to bring the worst possible behavior of the systems, we take the MAX(PIE\%) (highlighted in bold in Table 2 in the supplementary) as the raw score that is used to compute the final ratings shown in Table \ref{tab:ratings-pie}. Each of the values in the tuple in PIE \% shows the difference between APE before and after propensity score matching for different distributions. In each case, P0 is considered as the control group.

\noindent{\bf Interpretation:} From Table 2 in the supplementary, in most cases, choosing \textit{Industry} as the confounder led to more confounding bias in the systems. From Table \ref{tab:ratings-pie}, $S_{v2}$ and $S_{v1}$ exhibited the least amount of confounding bias with $S_{v2}$ outperforming all other systems in most cases. In most cases, systems exhibited more confounding bias when perturbations P1 and P2 were applied with P4 having the least impact and did not exacerbate the confounding bias significantly.  

\noindent{\bf Conclusion:} \textit{Industry} showed higher confounding bias. Both $S_{v1}$ and $S_{v2}$ performed better than all the systems, with $S_{v2}$ being the least biased of all. However, even they show some confounding bias. P1 and P2 proved to be the most impactful perturbations that led to more confounding bias. Hence, \emph{Confounder} affects the relationship between \emph{Perturbation} and \emph{Residual} of \emph{S} when \emph{Confounder} has an effect on \emph{Perturbation} which leads to confounding bias.

{\color{blue}\noindent{\bf Hypothesis-3:}} \emph{Perturbation} affects the \emph{Residual} of \emph{S} when \emph{Company} has an effect on \emph{Perturbation}.

\noindent{\bf Experimental Setup:}
The experimental setup in this experiment is same as that for Hypothesis-2. To compute the APE, we used PSM, the steps of which were detailed in the experimental setup of Hypothesis-2. PSM allows us to effectively determine the effect of \emph{Perturbation} on the \emph{Residual}. For instance, if two matched points belong to the same company but only one was perturbed, any difference in their residuals can be directly attributed to the perturbation itself rather than to other confounding factors. This method provides a clear understanding of the true impact of the \emph{Perturbation} on the \emph{Residual}.

\noindent{\bf Results:}
Table 2 in the supplementary shows the APE$_C$ \% (when \textit{Company} acts as the confounder) and APE$_I$ \% (when \textit{Industry} acts as the confounder) values computed for different systems across different distributions after applying PSM. As our rating method aims to bring the worst possible behavior of the systems, we take the MAX(APE) as the raw score (highlighted in bold in Table 2 in the supplementary) that is used to compute the final ratings. Each of the values in the tuple in APE shows the APE after propensity score matching across different distributions. In each case, P0 is considered as the control group.

\noindent{\bf Interpretation:} It is undesirable to have a higher APE, as it implies that the perturbation applied can have a significant impact on the residuals computed for different systems. From the APE values in Table 2 in the supplementary, when \textit{Industry} was considered as the confounder, it led to a higher APE. As the outcome of $S_b$ depended on the \textit{Company} (and varied from one company to another), the perturbation did not have any effect on the system. Whereas, when \textit{Industry} was considered as the confounder, the perturbation appeared influential, resulting in a high APE.  From Table 4 in the supplementary, perturbations had the least impact on $S_{v1}$ and $S_{v2}$ and more impact on $S_a$ (more than $S_r$). Among all the perturbations, P1 and P2 were the most disruptive.

\noindent{\bf Conclusion:} Perturbations had lesser impact on $S_{v1}$ and higher impact on $S_a$. P1 and P2 were the most disruptive perturbations. Considering \textit{Industry} as the confounder led to higher APE. Overall, \emph{Perturbation} affects the \emph{Residual} of \emph{S} when \emph{Company} has an effect on \emph{Perturbation}.

{\color{blue}\noindent{\bf Hypothesis-4:}} Semantic and compositional \emph{Perturbations} can degrade the performance of \emph{S}. 

\noindent{\bf Experimental Setup:} For this experiment, we do not use the causal model. The metric computation method is statistical. We compute the three accuracy metrics measured in \cite{zeng2023from} and summarized in Section \ref{sec:metrics}.

\begin{table}
\centering
   {\tiny
    \begin{tabular}{|l|p{0.4em}|p{5em}|p{5em}|p{5em}|p{2em}|p{5em}|}
    \hline
          {\bf Metric} &
          {\bf P} &    
          {\bf $S_{v1}$} &
          {\bf $S_{v2}$} &
          {\bf $S_a$} & 
          {\bf $S_b$} &
          {\bf $S_r$}
          \\ \hline 
          \multirow{6}{1em}{SMAPE$\downarrow$} & 
          P0 &
          \textbf{0.039 $\pm$ 0.035} &
           0.041 $\pm$ 0.037 &
           0.040 $\pm$ 0.037 &
          \multirow{6}{1.5em}{1.276 $\pm$ 0.663} &
          0.829 $\pm$ 0.638
          \\ \cline{2-5}
            \cline{7-7}
            & 
          P1 &
          \textbf{0.064 $\pm$ 0.167 }&
          0.127 $\pm$ 0.112 &
          0.084 $\pm$ 0.282 &
          &
          0.830 $\pm$ 0.639
          \\ \cline{2-5}
          \cline{7-7}
            & 
          P2 &
          \textbf{0.047 $\pm$ 0.069 }&
          0.068 $\pm$ 0.058 &
          0.069 $\pm$ 0.217 &
          &
          0.830 $\pm$ 0.639
          \\ \cline{2-5}
          \cline{7-7}
            & 
          P3 &
          \textbf{0.039 $\pm$ 0.035} &
          0.041 $\pm$ 0.037 &
          NA &
             &
          0.830 $\pm$ 0.640
          \\ \cline{2-5}
          \cline{7-7}
            & 
          P4 &
          \textbf{0.039 $\pm$ 0.035} &
          0.041 $\pm$ 0.037 &
          NA &
          &
          0.829 $\pm$ 0.639
          \\ \cline{2-5}
          \cline{7-7}
            & 
          P5 &
          0.089 $\pm$ 0.074 &
          \textbf{0.041 $\pm$ 0.035}  &
          NA &
          &
          0.832 $\pm$ 0.635
          \\ \hline

          \multirow{6}{1em}{MASE$\downarrow$} & 
          P0 &
          \textbf{3.68 $\pm$ 3.44} &
          3.89 $\pm$ 3.64 &
          3.79 $\pm$ 3.59 &
          \multirow{6}{1em}{947.56 $\pm$ 767.65} &
          86.45 $\pm$ 72.72
          \\ \cline{2-5}
          \cline{7-7}
             & 
          P1 &
          \textbf{5.30 $\pm$ 9.60} &
          11.18 $\pm$ 9.42 &
          18.36 $\pm$ 168.82 &
          &
          86.99 $\pm$ 73.53 
          \\ \cline{2-5}
          \cline{7-7}
             & 
          P2 &
          \textbf{4.24 $\pm$ 5.56} &
          6.16 $\pm$ 5.16 &
          8.24 $\pm$ 48.58 &
          &
          86.87 $\pm$ 73.32 
          \\ \cline{2-5}
          \cline{7-7}
             & 
          P3 &
          \textbf{3.68 $\pm$ 3.44} &
          3.89 $\pm$ 3.64 &
          NA &
          &
          86.65 $\pm$ 73.11 
          \\ \cline{2-5}
          \cline{7-7}
             & 
          P4 &
          \textbf{3.67 $\pm$ 3.43}   &
          3.90 $\pm$ 3.65 &
          NA &
          &
          86.52 $\pm$ 73.07 
          \\ \cline{2-5}
          \cline{7-7}
             & 
          P5 &
          8.26 $\pm$ 7.18 &
          \textbf{3.93 $\pm$ 3.54} &
          NA &
          &
          87.20 $\pm$ 73.14 
          \\ \hline
          \multirow{6}{2em}{Sign Accuracy (\%)$\uparrow$}   & 
          P0 &
          51.32 &
          51.28 &
          \textbf{60.08} &
          \multirow{6}{1em}{62.60} &
          49.88
          \\ \cline{2-5}
             & 
          P1 &
          48.77 &
          41.54 &
          \textbf{57.08} &
          &
          49.62
          \\\cline{2-5}
             & 
          P2 &
          \textbf{58.69} &
          45.28 &
          57.13 &
          &
          49.64
          \\ \cline{2-5}
              & 
          P3 &
          51.35 &
          \textbf{54.74} &
          NA &
          &
          49.71
          \\ \cline{2-5}
             & 
          P4 &
          \textbf{53.95} &
          51.14 &
          NA &
          &
          49.67
          \\\cline{2-5}
             & 
          P5 &
          43.97 &
          \textbf{52} &
          NA &
          &
          50.05
          \\ \hline
    \end{tabular}
    \caption{Performance metrics for test systems across different perturbations. SMAPE and MASE scores are reported as mean $\pm$ standard deviation. As P3 through P5 could not be applied to $S_a$, the metric values are marked as `NA'.}
    \label{tab:h4}
    }
\end{table}
\noindent{\bf Results:}
Table \ref{tab:h4} shows the performance metrics (used in \cite{zeng2023from}):
SMAPE, MASE, and Sign accuracy computed for all the test systems across all perturbations. As P3 through P5 could not be applied to $S_a$, we marked the metric values as `NA'. 

\noindent{\bf Interpretation:} For perturbations P1 through P4, $S_{v1}$ outperformed all other systems in terms of SMAPE and MASE. As $S_b$'s residuals vary depending on the ground truth stock prices of specific companies, $S_b$ consistently predicts the correct directional movement of stock prices. After $S_b$ (62.60 \%), $S_a$ shows higher sign accuracy for P0 and P1. For perturbations P2 through P5, the sign accuracies of $S_{v1}$ and $S_{v2}$ do not vary much compared to P0 (sometimes the accuracy exceeds P0). SMAPE and MASE values indicate that the performance of the models degrade when perturbations P1, P2 and P5 are applied compared to P3 and P4.

\noindent{\bf Conclusion:} Hence, semantic and compositional \emph{Perturbations} can degrade the performance of \emph{S}. 
\section{Discussion}
Our paper aimed to measure the impact of perturbations and confounders on the outcome of MM-TSFM.  We used the stock prices of six leading companies across three industries for our evaluation. 
However, we would like to collect historical data for more than two companies in each industry as future work. The analyses we did in the paper were with respect to \textit{Industry} and \textit{Company}. The motivation behind choosing these specific confounders is based on our intuition that, as stakeholders like investors and analysts are relying more on learning-based systems to decide whether to purchase stocks from a company, they would be interested to know if the errors committed by these learning-based models depend on the range of stock prices (which could serve as a proxy for the industry or company). For example, is a model committing more errors when predicting the stock prices for META compared to MRK? Errors can be committed by other factors such as the volatility of stock prices within an industry. To minimize the effect of volatility, we also performed intra-industry along with inter-industry analysis. As a part of future work, we would also like to do causal analysis considering other confounders such as seasonal trends and additional external factors also be done using additional sources such as financial news. 

As transformer-based approaches and especially multi-modal transformer-based approaches have been proven to be more efficient and accurate compared to traditional uni-modal neural network-based architectures \cite{liu2023survey}, we tested the robustness of one such model that was proved to be more effective for stock price prediction compared to other approaches \cite{zeng2023from}. 

The metrics we have chosen perfectly answered the research questions stated in Section \ref{sec:problem}. Instead of relying on just statistical accuracy metrics as the only performance metrics, we believe that one should select the metrics based on the questions one wants to answer. The hypothesis testing approach introduced in \cite{kausik2024rating}, which we have adapted for our work not only helped us answer the research questions but also quantified the biases and perturbation impacts on the outcome of the test systems. Traditional explainability approaches or other statistical approaches take control from human decision-makers, limiting their involvement \cite{miller2023explainable}. \cite{miller2023explainable} argues that hypothesis-driven decision support, though increases human cognitive load, can truly improve AI-assisted decision support. These approaches usually follow the process of abductive reasoning which involves forming hypotheses and judging their likelihood for explaining the systems' behavior to the end-users. Similarly, in our work, though we initially built a causal model based on our intuition, we tested the validity of each causal link by forming hypotheses and verifying their plausibility through the computation of different metrics. 

Each of the perturbations we used for our analysis has a real-world impact. Some of these could be applied to just numeric data and some can be applied to multi-modal data.  Methods like differential evaluation can be used to find the most impactful variation of a perturbation. But we only aimed to assess whether simple, and sometimes subtle perturbations would have an impact on the outcome of a time-series forecasting model. 



\section{Conclusion}
Our study introduces a causal analysis-based method for rating the robustness of Multimodal Time-Series Forecasting Models (MM-TSFM). We extend perturbation techniques to the MM-TSFM domain and evaluate their impact across six different perturbation settings, across industry and company. Notably, our findings highlight the disruptive nature of semantic perturbations, emphasizing the need for robustness evaluations beyond just accuracy metrics. We also find that the variant of ViT-num-spec model trained on a smaller but recent dataset is more robust against compositional perturbation compared to its larger variant. We also find that considering \textit{Industry} as the sensitive attribute / confounder led to an increase in bias compared to \textit{Company}. Moreover, our evaluation shows the commendable performance of ViT-num-spec models, which not only exhibited high accuracy but also demonstrated notable resilience across diverse experimental conditions compared to a traditional numerical baseline. Thus, within the scope of our experiments, multi-
modal (numeric+visual) forecasting exhibit greater robustness compared to simply numeric forecasting. 
Our rating system provides a practical tool for stakeholders to select robust solutions tailored to their specific data and operational requirements. 

\noindent {\textbf {Acknowledgements.}} This paper was prepared for informational purposes in part by the Artificial Intelligence Research group of JPMorgan Chase \& Co and its affiliates (“J.P. Morgan”) and is not a product of the Research Department of J.P. Morgan.  J.P. Morgan makes no representation and warranty whatsoever and disclaims all liability, for the completeness, accuracy or reliability of the information contained herein.  This document is not intended as investment research or investment advice, or a recommendation, offer or solicitation for the purchase or sale of any security, financial instrument, financial product or service, or to be used in any way for evaluating the merits of participating in any transaction, and shall not constitute a solicitation under any jurisdiction or to any person, if such solicitation under such jurisdiction or to such person would be unlawful.
\clearpage
\bibliography{references}


\end{document}


\maketitle


In this supplementary material, we provide the modified algorithms adapted from \cite{kausik2024rating} rate MM-TSFM for robustness. We also provided additional experimental results to help better understand our work in detail. The material is organized as follows:

\tableofcontents 

\section{Rating Algorithm}
We modified the rating algorithm introduced in \cite{kausik2024rating} to suit the MM-TSFM forecasting setting. Here is how the rating method works.
\begin{enumerate}
    \item Algorithm 1 computes the weighted rejection score (WRS) which was defined in equation 2 in the main paper.
    \item Algorithm 2 computes the PIE \% based on Propensity Score Matching (PSM) which was defined in equation 4 in the main paper.
    \item Algorithm 3 creates a partial order of systems within each perturbation based on the raw scores computed. It will arrange the systems in ascending order w.r.t the raw score. The final partial order (PO) will be a dictionary of dictionaries. 
    \item Algorithm 4 computes the final ratings for systems within each perturbation based on the PO from previous algorithm. It splits the set of raw score values obtained within each perturbation into `L' parts where `L' is the rating level chosen by the user. Each of the systems is given a rating based on the compartment number in which its raw score belongs. The algorithm will return a dictionary with perturbations as keys and ratings provided to each system within the perturbation as the value.
\end{enumerate}

\SetKwComment{Comment}{/* }{ */}
\begin{algorithm}
\small
	\caption{\emph{WeightedRejectionScore}}
	
	\textbf{Purpose:} is used to calculate the weighted sum of the number of rejections of null-hypothesis for Dataset $d_j$ pertaining to a system $s$, Confidence Intervals (CI) $ci_k$ and Weights $w_k$.
	
	\textbf{Input:}\\
	 $d$, dataset corresponding to a specific perturbation.\\
	 $CI$, confidence intervals (95\%, 70\%, 60\%). \\
      $s$, system for which WRS is being computed.   \\
     $W$, weights corresponding to different CIs (1, 0.8, 0.6).  \\
    \textbf{Output:} \\
     $Z$, Sensitive attribute  \\ 
	$\psi$, weighted rejection score.
	    $ \psi \gets 0$  
		\For {each $ci_i, w_i \in CI, W$} {
                    // $z_m, z_n$ are classes of $Z$

		            \For {each $z_m, z_n \in Z$} { 
		                $t, pval, dof \gets T-Test(z_m, z_n) $\; 
		                $t_{crit} \gets LookUp(ci_i, dof) $\;
		                \eIf{$t_{crit} > t$} {
		                    $\psi \gets \psi + 0 $\;
		                 }
		                    {$\psi \gets \psi + w_i $}
		               
		            }
    }
    \Return $\psi$
\end{algorithm}

\begin{algorithm}
\small
	\caption{ \emph{ComputePIEScore}}
	
	\textbf{Purpose:}  is used to calculate the Deconfounding Impact Estimation using Propensity Score Matching (PIE).
	
	\textbf{Input:}\\
	$s$, a system belonging to the set of test systems, $S$.\\
	$D$, datasets pertaining to a perturbation (different distributions).\\
        $p$, A perturbation other than $p_0$ \\
        $p_0$, control perturbation (or no perturbation. \\

    \textbf{Output:} \\
	$\psi$, PIE score.

    $ \psi \gets 0$
    
    $PIE\_list \gets []$
    \hspace{0.5 in} //  To store the list of PIE \% of all the datasets.

	\For {each $d_j \in D$} {
    	    $ APE\_o \gets  E(R|P=p_m) - E(R|P=p_0) $\; 
    	    $ APE\_m \gets  E(R|do(P=p_m)) - E(R|do(P=p_0)) $\; 
    		$PIE\_list[j] \gets (APE\_m - APE\_o) * 100$\;
        }
        $\psi \gets MAX(PIE\_list)$\;
    \Return $\psi$
\end{algorithm}

\begin{algorithm}
\small
	\caption{ \emph{CreatePartialOrder}}
	
	\textbf{Purpose:} is used to create a partial order based on the computed weighted rejection score / the PIE \%.
		
	\textbf{Input:}\\
	$S$, Set of systems.\\
        $P$, Set of perturbations.\\
    $F$, Flag that says whether the confounder is present (1) or not (0).  \\
    $D$, $CI$, $W$ (as defined in the previous algorithms). \\
    \textbf{Output:} \\
	$PO$, dictionary with a partial order for each perturbation.

	    $ PO \gets \{\}$\; 
            $SD \gets \{\}$;

	    \eIf {F == 0}{
		    \For {each $p_i \in P$}{ 
                    \For {each $s_j \in S$} {
		              $\psi \gets WeightedRejectionScore(p_i, s_j, D, CI, W)$\; 
		              $SD[s_j] \gets \psi$\;
 		    }
                $PO[p_i] \gets SORT(SD)$\;
                }
 		    }
 		    {\For {each $p_j \in P$} {
                    \For {each $s_i \in S$}{ 
		        $\psi \gets ComputePIEScore(s_i, p_j, p_0, D, CI, W)$\; 
		          $SD[s_j] \gets \psi$\;
 		    }
                $PO[p_i] \gets SORT(SD)$\;
                }
 		   }
	    \Return $PO$
\end{algorithm}

\begin{algorithm}
\small
	\caption{\emph{AssignRating}}
	
		\textbf{Purpose:} \emph{AssignRating} is used to assign a rating to each of the SASs based on the partial order and the number of rating levels, $L$.
		
	\textbf{Input:}\\
 $S$, $D$, $CI$, $W$, $P$ (as defined in the previous algorithms).
 
	$L$, rating levels chosen by the user.\\

    \textbf{Output:} \\
	$R$, dictionary with perturbations as keys and rating provided to each system within each perturbation as the value.

	    $ R \gets \{\} $\;
	    $ PO \gets  CreatePartialOrder(S,D,CI,W,G)$\;

        \For {$p_i$ \in P}{
        $ \psi \gets [ PO[p_i].values()]$\;

        \eIf {len(S) $>$ 1}{
 		$G \gets ArraySplit(\psi, L) $\;
 		\For {$k,i \in PO[p_i],\psi $}{
 		        \If {$i \in g_j$}{
 		            $SD[k] \gets j$\;
 		        }
 		       }
 	    }
 	    { 	    //  Case of single SAS in $S$
 	
 	\eIf {$\psi$ $==$ 0} {$SD[k] \gets 1$ \hspace{0.5 in}  } 
 	    {$SD[k] \gets $L$ $  \hspace{0.5 in}  }
 	    } 
      $R[p_i] \gets SD$
      }
		\Return $R$
\end{algorithm}

\section{More Experimental Results}
\begin{table}[htb]
\centering
   {\small
    \begin{tabular}{|p{2.5em}|p{1em}|c|c|c|c|c|c|}
    \cline{3-8}  
          \multicolumn{1}{c}{}&
          \multicolumn{1}{c|}{}&
          \multicolumn{3}{c|}{\bf Inter-industry} &
          \multicolumn{3}{c|}{\bf Intra-industry}
          \\ \hline

          {\bf System}     &  
          {\bf P}      &  
          {\bf $Z_t$$Z_p$} &
          {\bf $Z_t$$Z_f$} &
          {\bf $Z_p$$Z_f$} &
          {\bf $Z_m$$Z_g$} &
          {\bf $Z_{pf}$$Z_{mr}$} &
          {\bf $Z_w$$Z_c$} 
          \\ \hline  
          
          \multirow{6}{2em}{$S_{v1}$} & 
          P0 &
          15.28$^1$  &
          16.25$^1$ &
          2.04$^2$ &
          13.13$^1$ &
          16.58$^1$ &
          1.73$^2$
         \\ \cline{2-8}
          & 
          P1 &
          10.26$^1$&
          10.90$^1$&
          2.01$^2$ &
          7.79$^1$ &
          7.07$^1$ &
          0.85
         \\ \cline{2-8}
         & 
          P2 &
          13.69$^1$ &
          14.14$^1$ &
          1.51$^2$  &
          10.17$^1$ &
          10.72$^1$ &
          0.33
         \\ \cline{2-8}
          & 
          P3 &
          15.82$^1$  &
          16.25$^1$ &
          2.04$^2$ &
          13.13$^1$ &
          16.58$^1$ &
          1.73$^2$
         \\ \cline{2-8}
          &
          P4 &
          16.02$^1$  &
          16.45$^1$ &
          2.02$^2$ &
          13.24$^1$ &
          16.60$^1$ &
          1.43$^3$
         \\ \cline{2-8}
          &
          P5 &
          16.34$^1$  &
          16.03$^1$ &
          2.03$^2$ &
          15.93$^1$ &
          20.85$^1$ &
          5.70$^1$
         \\ \hline
          
          \multirow{6}{2em}{$S_{v2}$} & 
          P0 &
          15.97$^1$ &
          16.22$^1$ &
          0.73 &
          13.56$^1$ &
          16.37$^1$ &
          0.59
         \\ \cline{2-8}
          & 
          P1 &
          18.67$^1$ &
          20.82$^1$ &
          4.9$^1$ &
          13.40$^1$ &
          16.03$^1$ &
          0.66
         \\ \cline{2-8}
         & 
          P2 &
          17.62$^1$ &
          18.78$^1$ &
          4.84$^1$ &
          11.54$^1$ &
          15.36$^1$ &
          1.34$^3$
         \\ \cline{2-8}
          & 
          P3 &
          15.97$^1$ &
          16.22$^1$ &
          0.74 &
          13.57$^1$ &
          16.37$^1$ &
          0.59
         \\ \cline{2-8}
          &
          P4 &
          16.02$^1$ &
          16.28$^1$ &
          0.74 &
          13.60$^1$ &
          16.43$^1$ &
          0.54
         \\ \cline{2-8}
          &
          P5 &
          15.43$^1$ &
          15.77$^1$ &
          1.41$^3$ &
          13.44$^1$ &
          16.63$^1$ &
          1.73$^2$
         \\ \hline
        \multirow{3}{2em}{$S_a$} & 
          P0 &
          16.86$^1$ &
          17.35$^1$ &
          1.66$^2$ &
          14.43$^1$ &
          16.23$^1$ &
          3.54$^1$
         \\ \cline{2-8}
          & 
          P1 &
          1.48$^2$ &
          2.14$^2$ &
          1.24     &
          1.43$^3$ &
          1.22     &
          0.58
         \\ \cline{2-8}
             & 
          P2 &
          2.74$^1$ &
          3.41$^1$ &
          1.07     &
          2.02$^2$ &
          1.59$^2$     &
          0.55
         \\ \hline
          
          \multirow{6}{2em}{$S_b$} &
          P0 &
          H$^1$ &
          H$^1$ &
          H$^1$ &
          H$^1$ &
          H$^1$ &
          H$^1$
         \\ \cline{2-8}
          & 
          P1 &
          H$^1$ &
          H$^1$ &
          H$^1$ &
          H$^1$ &
          H$^1$ &
          H$^1$
         \\ \cline{2-8}
         & 
          P2 &
          H$^1$ &
          H$^1$ &
          H$^1$ &
          H$^1$ &
          H$^1$ &
          H$^1$
         \\ \cline{2-8}
          & 
          P3 &
          H$^1$ &
          H$^1$ &
          H$^1$ &
          H$^1$ &
          H$^1$ &
          H$^1$
         \\ \cline{2-8}
          &
          P4    &
          H$^1$ &
          H$^1$ &
          H$^1$ &
          H$^1$ &
          H$^1$ &
          H$^1$
         \\ \cline{2-8}
          &
          P5 &
          H$^1$ &
          H$^1$ &
          H$^1$ &
          H$^1$ &
          H$^1$ &
          H$^1$
         \\ \hline
          \multirow{6}{2em}{$S_r$} &
          P0 &
          20.22$^1$ &
          20.34$^1$ &
          0.43      &
          40.91$^1$ &
          15.61$^1$ &
          0.38      
         \\ \cline{2-8}
          & 
          P1 &
          20.72$^1$ &
          20.61$^1$ &
          1.04      &
          43.47$^1$ &
          13.51$^1$ &
          2.16$^2$      
         \\ \cline{2-8}
         & 
          P2 &
          20.51$^1$ &
          20.75$^1$ &
          1.71$^2$  &
          44.94$^1$ &
          15.04$^1$ &
          1.27      
         \\ \cline{2-8}
          & 
          P3 &
          20.84$^1$ &
          20.76$^1$ &
          1.02      &
          44.82$^1$ &
          13.10$^1$ &
          2.29$^1$      
         \\ \cline{2-8}
          &
          P4 &
          20.53$^1$ &
          20.76$^1$ &
          1.39$^3$  &
          41.35$^1$ &
          15.50$^1$ &
          1.92$^2$      
         \\ \cline{2-8}
          &
          P5 &
          20.62$^1$ &
          20.68$^1$ &
          0.26      &
          43.46$^1$ &
          13.04$^1$ &
          1.05      
         \\ \hline
    \end{tabular}
    \caption{Table showing the t-values across different systems and perturbation groups when there is no causal link from \emph{Perturbation} to \emph{Residual}. The superscript shows whether the null hypothesis is rejected or accepted in each case for the CIs considered (95\%, 70\%, 60\%). `1' indicates rejection with all 3 CIs, `2' indicates rejection with all both 70\% and 60\% CIs, `3' indicates rejection with 60\%. Each $Z_iZ_j$ denote the t-value computed between the distributions, $(R^t_{max} | Z_i)$ and $(R^t_{max} | Z_j)$.}
    \label{tab:h1}
    }
\end{table}

\begin{table*}[htb]
\centering
   {\tiny
    \begin{tabular}
    {|p{2.5em}|p{0.5em}|p{2.5em}|p{2.5em}|p{12em}|p{16em}|p{11em}|p{16em}|}
        \hline
          {\bf System} &  
          {\bf P} &  
          {\bf WRS$_{I}$} &
          {\bf WRS$_{C}$} &
          {\bf MAX(PIE$_{C}$) \%} &
          {\bf MAX(PIE$_{I}$) \%} &
          {\bf MAX(ATE$_{C}$) \%} &
          {\bf MAX(ATE$_{I}$) \%} 
          \\ \hline  
          
          \multirow{9}{2em}{$S_{v1}$} &
          P0  &
          5.9 &
          5.9 &
          NA &
          NA &
          NA &
          NA
          \\ \cline{2-8}
             &
          P1 &
          \textbf{5.9}   &
          4.6 &
          (278.69, 117.07, \textbf{630.10}) &
          (\textbf{551}, 395.45, 229.32, 233.16, 156.51, 466.08) &
          (2.79, 1.38, \textbf{6.53}) &
          (\textbf{6.05}, 5.47, 2.89, 2.66, 1.77, 4.69)
          \\ \cline{2-8}
             & 
          P2 &
          \textbf{5.9}&
          4.6 &
          (471.89, \textbf{941.93}, 277.72) &
          (86.15, 68.59, 35.92, \textbf{898.90}, 107.24, 662.79) &
          (6.16, \textbf{10.97}, 3.47)&
          (0.97, 2.31, 0.79, \textbf{10.10}, 1.51, 7.77)
          \\\cline{2-8}
             & 
          P3 &
          5.9&
          5.9&
          (176.33, \textbf{276.86}, 113.29)&
          (\textbf{2427.35}, 108.79, 123.73, 84.41, 397.51, 156.08)&
          (3.49, \textbf{4.15}, 3.18)&
          (\textbf{25.53}, 1.19, 0.98, 3.35, 5.84, 3.04)
          \\\cline{2-8}
          & 
          P4 &
          \textbf{5.9}&
          5.2 &
          (\textbf{229.03}, 140.12, 33.44)&
          (213.98, 115.61, 137.44, 183.56, \textbf{247.80}, 65.19)&
          (\textbf{4.22}, 0.48, 2.27)&
          (4.21, 3.11, 0.40, 0.86, \textbf{4.98}, 1.64)
          \\\cline{2-8}
          & 
          P5 &
          5.9&
          \textbf{6.9} &
          (\textbf{344}, 18.70, 155.73)&
          (358.31, 68.18, 83.95, 102.42, 48.66, \textbf{378.19})&
          (\textbf{13.20}, 5.53, 7.66)&
          (\textbf{14.02}, 7.50, 6.24, 6.29, 8.32, 9.76)
          \\\hline
          \multirow{9}{2em}{$S_{v2}$} & 
          P0 &
          4.6&
          4.6 &
          NA &
          NA &
          NA &
          NA
          \\ \cline{2-8}
             &
         P1  &
         \textbf{6.9} &
         4.6        &     
         (\textbf{1191.91}, 253.75, 571.03)&
          (382.22, \textbf{415.74}, 84.41, 303.31, 4.08, 355.39) &
          (2.66, 11.40, \textbf{13.93})&
          (\textbf{18.29}, 15.27, 9.07, 13.48, 8.81, 12.80)
          \\ \cline{2-8}
          & 
          P2 &
          \textbf{6.9}&
          5.2 &
          (186.95, \textbf{1490.65}, 434.33)&
          (\textbf{575.12}, 250.40, 201.10, 150.92, 146.44, 267)&
          (4.07, \textbf{15.82}, 5.58)&
          (\textbf{6.42}, 3.02, 2.60, 1.66, 1.59, 3.24)
          \\ \cline{2-8}
          & 
          P3 &
          4.6&
          4.6&
          (140.11, 8.88, \textbf{224.98})&
          (38.21, 145, 169.90, 256.89, 73.86, \textbf{1277.44})&
          (\textbf{4.90}, 2.79, 4.61)&
          (3.98, 1.94, 0.60, 0, 3.60, \textbf{15.75})
          \\ \cline{2-8}
          & 
          P4 &
          4.6&
          4.6&
          (1222.16, \textbf{1694.57}, 59.37)&
          (16.71, 128.47, 143.78, \textbf{942.02}, 107.62, 217.33)&
          (15.43, \textbf{19.93}, 3.30)&
          (1.73, 4.16, 1.33, \textbf{12.18}, 0.81, 0.58)
          \\\cline{2-8}
          & 
          P5 &
          5.2&
          \textbf{5.9}&
          (\textbf{273.12}, 239.87, 180.94)&
          (184.42, 64.65, 172.05, \textbf{284.95}, 64.08, 25.94)&
          (1.27, \textbf{4.94}, 0.94)&
          (1.09, 1.67, 0.91, 0.57, \textbf{3.80}, 2.72)
          \\ \hline
          \multirow{4}{2em}{$S_a$} & 
          P0 &
          5.9&
          \textbf{6.9}&
          NA &
          NA &
          NA &
          NA
          \\\cline{2-8}
             &
          P1 &
          \textbf{2.6}&
          0.6&
          (209.62, \textbf{982.38}, 815.15) &
          (763.20, 686.91, 483.04, 836.27, 416.46, \textbf{914.64})&
          (40.93, \textbf{61.87}, 37.35) &
          (\textbf{59.80}, 27.69, 33.58, 44.08, 37.77, 24.10)
          \\ \cline{2-8}
             & 
          P2 &
          \textbf{4.6}&
          2.6  & 
          (509.86, \textbf{1275.04}, 253.82) &
          (1121.43, 301.48, 380.24, 787.47, \textbf{1154.87})&
          (10.51, \textbf{11.32}, 10.76)&
          (\textbf{21.39}, 9.63, 9.38, 7.05, 5.07, 9.51)
          \\ \hline
          \multirow{9}{2em}{$S_b$} & 
          P0 &
          6.9&
          6.9&
          NA &
          NA &
          NA &
          NA
         \\ \cline{2-8}
             & 
          P1 &
          6.9&
          6.9&
          (\textbf{6916.11}, 3919.36, 3508.78)&
          (27.08, 1513.58, 478.55, 69.47, 984.96, \textbf{3283.88})&
          (\textbf{101.31}, 46.19, 51.81)&
          (0, 0, 0, 0, 0, 0)
         \\ \cline{2-8}
             &
          P2 &
          6.9&
          6.9&
          (\textbf{9474.61}, 647.83, 2850.52)&
          (2109.78, 1312.08, 1468.52, 1181.40, \textbf{2174.39}, 1223.53)&
          (\textbf{101.20}, 9.86, 54.54)&
          (0, 0, 0, 0, 0, 0)
         \\ \cline{2-8}
             &
          P3 &
          6.9&
          6.9&
          (3084.27, 3876.09, \textbf{7489.48})&
          (1208.21, 528.82, 620.54, \textbf{1846.56}, 902.16, 750.90) &
          (50.98, 47.53, \textbf{99.72})&
          (0, 0, 0, 0, 0, 0) 
         \\ \cline{2-8}
             &
          P4 &
          6.9&
          6.9&
          (1807.07, 4245.41, \textbf{7618.25})&
          (\textbf{3557.45}, 751.37, 252, 929.87, 1117.87, 2707.19) &
          (5.07, 51.11, \textbf{100.20})&
          (0, 0, 0, 0, 0, 0)  
          \\ \cline{2-8}
             &
          P5 &
          6.9&
          6.9&
          (4102.07, \textbf{8966.57}, 3762.97)&
          (\textbf{2118.88}, 973.35, 58.66, 581.74, 1233.95, 547.38)&
          (49.40, \textbf{98.61}, 42.85)&
          (0, 0, 0, 0, 0, 0)   
          \\ \hline
          \multirow{9}{2em}{$S_r$} & 
          P0 &
          4.6&
          4.6&
          NA &
          NA &
          NA &
          NA
          \\ \cline{2-8}
             & 
          P1 &
          4.6&
          \textbf{5.9}&
          (\textbf{4756.40}, 983.79, 2262.36)&
          (1019.20, 297.76, 657.79, \textbf{1041.01}, 308.84, 102.45)&
          (\textbf{48.80}, 10.10, 23.11)&
          (\textbf{15.36}, 4.63, 7.93, 14, 3.84, 2.81)
          \\ \cline{2-8}
             & 
          P2 &
          \textbf{5.9} &
          4.6&
          (\textbf{4274.38}, 1338.71, 1527.33)&
          (177.63, 51.40, \textbf{1463.71}, 183.58, 70.95, 7.15)&
          (\textbf{42.91}, 16.62, 21.16)&
          (7.90, 1.30, \textbf{17.61}, 0.43, 3.35, 10.15)    
          \\ \cline{2-8}
             & 
          P3 & 
          4.6&
          \textbf{6.9}&
          (\textbf{3560.94}, 1648.90, 2193.67)&
          (43.53, 135.03, 912.42, 558.95, \textbf{1305.78}, 430.51)&
          (\textbf{36.59}, 17.39, 24.88)&
          (2.28, 1.65, 11.27, 6.23, \textbf{16.63}, 7.21)  
          \\ \cline{2-8}
             & 
          P4 &
          5.2 &
          \textbf{5.9}&
          (1922.55, 1523.92, \textbf{2250.35})&
          (575.48, 386.15, 97.74, 83.20, 67.56, \textbf{1314.82})&
          (23.66, 17.32, \textbf{23.75})&
          (9.20, 4.87, 1.76, 1.56, 2.81, \textbf{15.18})     
          \\ \cline{2-8}
               & 
            P5 &  
            4.6 &
            4.6 &
          (\textbf{4025.31}, 2843.16, 2401.06)&
          (\textbf{1928.21}, 1703.29, 247.17, 360.41, 381.82, 516.29)  &
          (\textbf{44.11}, 28.63, 27)&
          (\textbf{21.44}, 17.11, 5.69, 6.16, 4.52, 8.50)
          \\ \hline
    \end{tabular}
    \caption{Table showing the Weighted Rejection Scores (WRS) across companies within the same industry (WRS$_{Industry}$) and across different industries (WRS$_{Company}$), the PIE \% values computed for different systems across different distributions with each value in the tuple showing the difference between ATE before and after propensity score matching for different perturbation groups (P1 through P5),
    and the APE scores with $Industry$ as the confounder and $Company$ as the confounder.}
    \label{tab:raw-scores}
    }
\end{table*}

\begin{table*}[!h]
\centering
{\small
    \begin{tabular}{|p{8em}|p{0.8em}|p{14em}|p{14em}|}
    \hline
          {\bf Forecasting Evaluation Dimensions} &
          {\bf P} &    
          {\bf Partial Order} &
          {\bf Complete Order} 
          \\ \hline 
          \multirow{5}{6em}{Inter-industry statistical bias (WRS$_I$)} &
          P0 & 
          \{$S_{v2}$: 4.6, $S_r$: 4.6, $S_{v1}$: 5.9, $S_a$: 5.9, $S_b$: 6.9 \} &
          \{\textbf{$S_{v2}$: 1}, $S_r$: 1, $S_{v1}$: 2, $S_a$: 2, $S_b$: 3 \}
          \\ \cline{2-4}
          &
          P1 & 
          \{$S_a$: 2.6, $S_r$: 4.6, $S_{v1}$: 5.9, $S_{v2}$: 6.9, $S_b$: 6.9 \} &
          \{\textbf{$S_a$: 1}, $S_r$: 2, $S_{v1}$: 2, $S_{v2}$: 3, $S_b$: 3\}
          \\ \cline{2-4}
          &
          P2 & 
          \{$S_a$: 4.6, $S_r$: 4.6, $S_{v1}$: 5.9, $S_{v2}$: 6.9, $S_b$: 6.9\} &
          \{\textbf{$S_a$: 1}, $S_r$: 1, $S_{v1}$: 2, $S_{v2}$: 3, $S_b$: 3\}
          \\ \cline{2-4}
          &
          P3 & 
          \{$S_{v2}$: 4.6, $S_r$: 4.6, $S_{v1}$: 5.9, $S_b$: 6.9 \} &
          \{\textbf{$S_{v2}$: 1}, $S_r$: 1, $S_{v1}$: 2, $S_b$: 3\}
          \\ \cline{2-4}
          &
          P4 & 
          \{$S_{v2}$: 4.6, $S_r$: 5.2, $S_{v1}$: 5.9, $S_b$: 6.9\} &
          \{\textbf{$S_{v2}$: 1}, $S_r$: 2, $S_{v1}$: 2, $S_b$: 3\}
          \\ \cline{2-4}
          &
          P5 & 
          \{$S_r$: 4.6, $S_{v2}$: 5.2, $S_{v1}$: 5.9, $S_b$: 6.9\} &
          \{\textbf{$S_r$: 1}, $S_{v2}$: 2, $S_{v1}$: 2, $S_b$: 3\}
          \\ \hline
          \multirow{5}{6em}{Intra-industry statistical bias (WRS$_C$)} &
          P0 & 
          \{$S_{v2}$: 4.6, $S_r$: 4.6, $S_{v1}$: 5.9, $S_r$: 6.9, $S_b$: 6.9 \} &
          \{\textbf{$S_{v2}$: 1}, $S_r$: 1, $S_{v1}$: 2, $S_a$: 3, $S_b$: 3 \}
          \\ \cline{2-4}
          &
          P1 & 
          \{$S_a$: 0.6, $S_{v1}$: 4.6, $S_{v2}$: 4.6, $S_r$: 5.9, $S_b$: 6.9 \} &
          \{\textbf{$S_a$: 1}, $S_{v1}$: 1, $S_{v1}$: 1, $S_a$: 3, $S_b$: 3 \}
          \\ \cline{2-4}
          &
          P2 & 
          \{$S_a$: 2.6, $S_{v1}$: 4.6, $S_r$: 4.6, $S_{v2}$: 5.2, $S_b$: 6.9 \} &
          \{\textbf{$S_a$: 1}, $S_{v1}$: 1, $S_r$: 1, $S_a$: 2, $S_b$: 3 \}
          \\ \cline{2-4}
          &
          P3 & 
          \{$S_{v2}$: 4.6, $S_{v1}$: 5.9, $S_r$: 6.9, $S_b$: 6.9 \} &
          \{\textbf{$S_{v2}$: 1}, $S_r$: 2, $S_r$: 3, $S_b$: 3 \}
          \\ \cline{2-4}
          &
          P4 & 
          \{$S_{v2}$: 4.6, $S_{v1}$: 5.2, $S_r$: 5.9, $S_b$: 6.9 \} &
          \{\textbf{$S_{v2}$: 1}, $S_r$: 1, $S_r$: 2, $S_b$: 3 \}
          \\ \cline{2-4}
          &
          P5 & 
          \{$S_r$: 4.6, $S_{v2}$: 5.9, $S_{v1}$: 6.9, $S_b$: 6.9 \} &
          \{\textbf{$S_r$: 1}, $S_{v2}$: 2, $S_{v1}$: 3, $S_b$: 3 \}
          \\ \hline
          \multirow{5}{6em}{Accuracy (SMAPE)} &
          P0 & 
          \{$S_{v1}$: 0.039, $S_a$: 0.040, $S_{v2}$: 0.041, $S_r$: 0.829, $S_b$: 1.276 \} &
          \{\textbf{$S_{v1}$: 1}, $S_a$: 1, $S_{v2}$: 2, $S_r$: 2, $S_b$: 3 \} 
          \\ \cline{2-4}
          &
          P1 & 
          \{$S_{v1}$: 0.064, $S_a$: 0.084, $S_{v2}$: 0.127, $S_r$: 0.830, $S_b$: 1.276 \} &
          \{\textbf{$S_{v1}$: 1}, $S_a$: 1, $S_{v2}$: 2, $S_r$: 2, $S_b$: 3 \} 
          \\ \cline{2-4}
          &
          P2 & 
          \{$S_{v1}$: 0.047, $S_{v2}$: 0.068, $S_a$: 0.069, $S_r$: 0.830, $S_b$: 1.276 \} &
          \{\textbf{$S_{v1}$: 1}, $S_{v2}$: 1, $S_a$: 2, $S_r$: 2, $S_b$: 3 \} 
          \\ \cline{2-4}
          &
          P3 & 
          \{$S_{v1}$: 0.039, $S_{v2}$: 0.041, $S_r$: 0.830, $S_b$: 1.276 \} &
          \{\textbf{$S_{v1}$: 1}, $S_{v2}$: 1, $S_r$: 2, $S_b$: 3 \} 
          \\ \cline{2-4}
          &
          P4 & 
          \{$S_{v1}$: 0.039, $S_{v2}$: 0.041, $S_r$: 0.829, $S_b$: 1.276 \} &
          \{\textbf{$S_{v1}$: 1}, $S_{v2}$: 1, $S_r$: 2, $S_b$: 3 \} 
          \\ \cline{2-4}
          &
          P3 & 
          \{$S_{v2}$: 0.041, $S_{v1}$: 0.089, $S_r$: 0.832, $S_b$: 1.276 \} &
          \{\textbf{$S_{v2}$: 1}, $S_{v1}$: 1, $S_r$: 2, $S_b$: 3 \} 
          \\ \hline
    \end{tabular}
    }
    \caption{Table showing final raw scores and ratings based on WRS and SMAPE \%. Higher rating indicate higher statistical bias (or error, for SMAPE). For simplicity, we denoted the raw scores for SMAPE using just the mean value, but standard deviation was also considered for rating. The chosen rating level, L = 3.}
    \label{tab:ratings-wrs}

\end{table*}

\begin{table*}[!h]
\centering
{\small
    \begin{tabular}{|p{8em}|p{0.8em}|p{14em}|p{14em}|}
    \hline
          {\bf Forecasting Evaluation Dimensions} &
          {\bf P} &    
          {\bf Partial Order} &
          {\bf Complete Order} 
          \\ \hline 
          \multirow{5}{6em}{Perturbation Impact with \textit{Industry} as the confounder (APE$_I$)} &
          P1 & 
          \{$S_{v1}$: 6.53, $S_{v2}$: 13.93, $S_r$: 48.80, $S_a$: 61.87, $S_b$: 101.31 \} &
          \{\textbf{$S_{v1}$: 1}, $S_{v2}$: 1, $S_r$: 2, $S_a$: 3, $S_b$: 3 \}
          \\ \cline{2-4}
          &
          P2 & 
          \{$S_{v1}$: 10.97, $S_a$: 11.32, $S_{v2}$: 15.82, $S_r$: 42.91, $S_b$: 101.20 \} &
          \{\textbf{$S_{v1}$: 1}, $S_a$: 1, $S_{v2}$: 2, $S_r$: 3, $S_b$: 3 \}
          \\ \cline{2-4}
          &
          P3 & 
          \{$S_{v1}$: 4.15, $S_{v2}$: 4.90, $S_r$: 36.59, $S_b$: 99.72 \} &
          \{\textbf{$S_{v1}$: 1}, $S_{v2}$: 1, $S_r$: 2, $S_b$: 3 \}
          \\ \cline{2-4}
          &
          P4 & 
          \{$S_{v1}$: 4.22, $S_{v2}$: 19.93, $S_r$: 23.75, $S_b$: 100.20\} &
          \{\textbf{$S_{v1}$: 1}, $S_{v2}$: 1, $S_r$: 2, $S_b$: 3\}
          \\ \cline{2-4}
          &
          P5 & 
          \{$S_{v2}$: 4.94, $S_{v1}$: 13.20, $S_r$: 44.11, $S_b$: 98.61\} &
          \{\textbf{$S_{v2}$: 1}, $S_{v1}$: 1, $S_r$: 2, $S_b$: 3\}
          \\ \hline
          \multirow{5}{6em}{Perturbation Impact with \textit{Company} as the confounder (APE$_C$)} &
          P1 & 
          \{$S_b$: 0, $S_{v1}$: 6.05, $S_r$: 15.36, $S_{v2}$: 18.29, $S_a$: 59.80 \} &
          \{\textbf{$S_b$: 1}, $S_{v1}$: 1, $S_r$: 2, $S_{v2}$: 3, $S_a$: 3 \}
          \\ \cline{2-4}
          &
          P2 & 
          \{$S_b$: 0, $S_{v2}$: 6.42, $S_{v1}$: 10.10, $S_r$: 17.61, $S_a$: 21.39 \} &
          \{\textbf{$S_b$: 1}, $S_{v2}$: 1, $S_{v1}$: 2, $S_r$: 3,  $S_a$: 3 \}
          \\ \cline{2-4}
          &
          P3 & 
          \{$S_b$: 0, $S_{v2}$: 15.75, $S_r$: 16.63, $S_{v1}$: 25.53 \} &
          \{\textbf{$S_b$: 1}, $S_{v2}$: 1, $S_r$: 2, $S_{v1}$: 3 \}
          \\ \cline{2-4}
          &
          P4 & 
          \{$S_b$: 0, $S_{v1}$: 4.98, $S_{v2}$: 12.18, $S_r$: 15.18 \} &
          \{\textbf{$S_b$: 1}, $S_{v1}$: 1, $S_{v2}$: 2, $S_r$: 3 \}
          \\ \cline{2-4}
          &
          P5 & 
          \{$S_b$: 0, $S_{v2}$: 3.80, $S_{v1}$: 14.02, $S_r$: 21.44 \} &
          \{\textbf{$S_b$: 1}, $S_{v2}$: 1, $S_{v1}$: 2, $S_r$: 3 \}
          \\ \hline
          \multirow{5}{6em}{Accuracy (Sign Accuracy \%)} &
          P0 & 
          \{$S_r$: 49.88, $S_{v2}$: 51.28, $S_{v1}$: 51.32, $S_a$: 60.08, $S_b$: 62.60 \} &
          \{\textbf{$S_r$: 1}, $S_{v2}$: 1, $S_{v1}$: 2, $S_a$: 2, $S_b$: 3 \}          
          \\ \cline{2-4}
          &
          P1 & 
          \{$S_{v2}$: 41.54, $S_{v1}$: 48.77, $S_r$: 49.62, $S_a$: 57.08, $S_b$: 62.60 \} &
          \{\textbf{$S_{v2}$}: 1, $S_{v1}$: 1, $S_r$: 2, $S_a$: 2, $S_b$: 3 \} 
          \\ \cline{2-4}
          &
          P2 & 
          \{$S_{v2}$: 45.28, $S_r$: 49.64, $S_a$: 57.13, $S_{v1}$: 58.69, $S_b$: 62.60 \} &
          \{\textbf{$S_{v2}$}: 1, $S_r$: 1, $S_a$: 2, $S_{v1}$: 2, $S_b$: 3 \} 
          \\ \cline{2-4}
          &
          P3 & 
          \{$S_r$: 49.71, $S_{v1}$: 51.35, $S_{v2}$: 54.74, $S_b$: 62.60 \} &
          \{\textbf{$S_r$: 1}, $S_{v1}$: 1, $S_{v2}$: 2, $S_b$: 3 \} 
          \\ \cline{2-4}
          &
          P4 & 
          \{$S_r$: 49.67, $S_{v2}$: 51.14, $S_{v1}$: 53.95, $S_b$: 62.60 \} &
          \{\textbf{$S_r$: 1}, $S_{v2}$: 1, $S_{v1}$: 2, $S_b$: 3 \} 
          \\ \cline{2-4}
          &
          P5 & 
        \{$S_{v1}$: 43.97, $S_r$: 50.05, $S_{v2}$: 52, $S_b$: 62.60 \} &
        \{\textbf{$S_{v1}$}: 1, $S_r$: 1, $S_{v2}$: 2, $S_b$: 3 \} 
          \\ \hline
    \end{tabular}
    }
    \caption{Table showing final raw scores and ratings based on APE and Sign Accuracy \%. Higher rating indicate higher impact of perturbation and higher accuracy for sign accuracy. The chosen rating level, L = 3.}
    \label{tab:ratings-ape}
\end{table*}

\clearpage
\printbibliography